\documentclass[10pt,twocolumn,letterpaper]{article}

\usepackage[pagenumbers]{cvpr} 

\usepackage{booktabs}
\usepackage{multirow}
\usepackage{calc}
\usepackage{subcaption}
\usepackage[table]{xcolor} 
\usepackage{array}
\usepackage{stfloats}
\usepackage{verbatim}
\usepackage{xcolor}
\usepackage{caption}

\definecolor{cvprblue}{rgb}{0.21,0.49,0.74}
\usepackage[pagebackref,breaklinks,colorlinks,allcolors=cvprblue]{hyperref}

\def\paperID{2838} 
\def\confName{CVPR}
\def\confYear{2026}

\title{One-to-All Animation: Alignment-Free Character Animation \\
and Image Pose Transfer}

\author{
Shijun Shi$^{1*}$ \quad 
Jing Xu$^{2*}$ \quad 
Zhihang Li$^{3}$ \quad
Chunli Peng$^{4}$ \quad
Xiaoda Yang$^{5}$ \\
Lijing Lu$^{3}$ \quad
Kai Hu$^{1\dagger}$ \quad 
Jiangning Zhang$^{5\dagger}$  \\
$^{1}$Jiangnan University \quad
$^{2}$University of Science and Technology of China \\
$^{3}$Chinese Academy of Sciences \quad
$^{4}$Beijing University of Posts and Telecommunications \\
$^{5}$Zhejiang University \\
\\[-3mm]
{\small\url{https://ssj9596.github.io/one-to-all-animation-project/}}\\
}

\begin{document}
\twocolumn[{%
\maketitle
\vspace{-1.2cm}
\begin{center}
    \includegraphics[width=\textwidth]{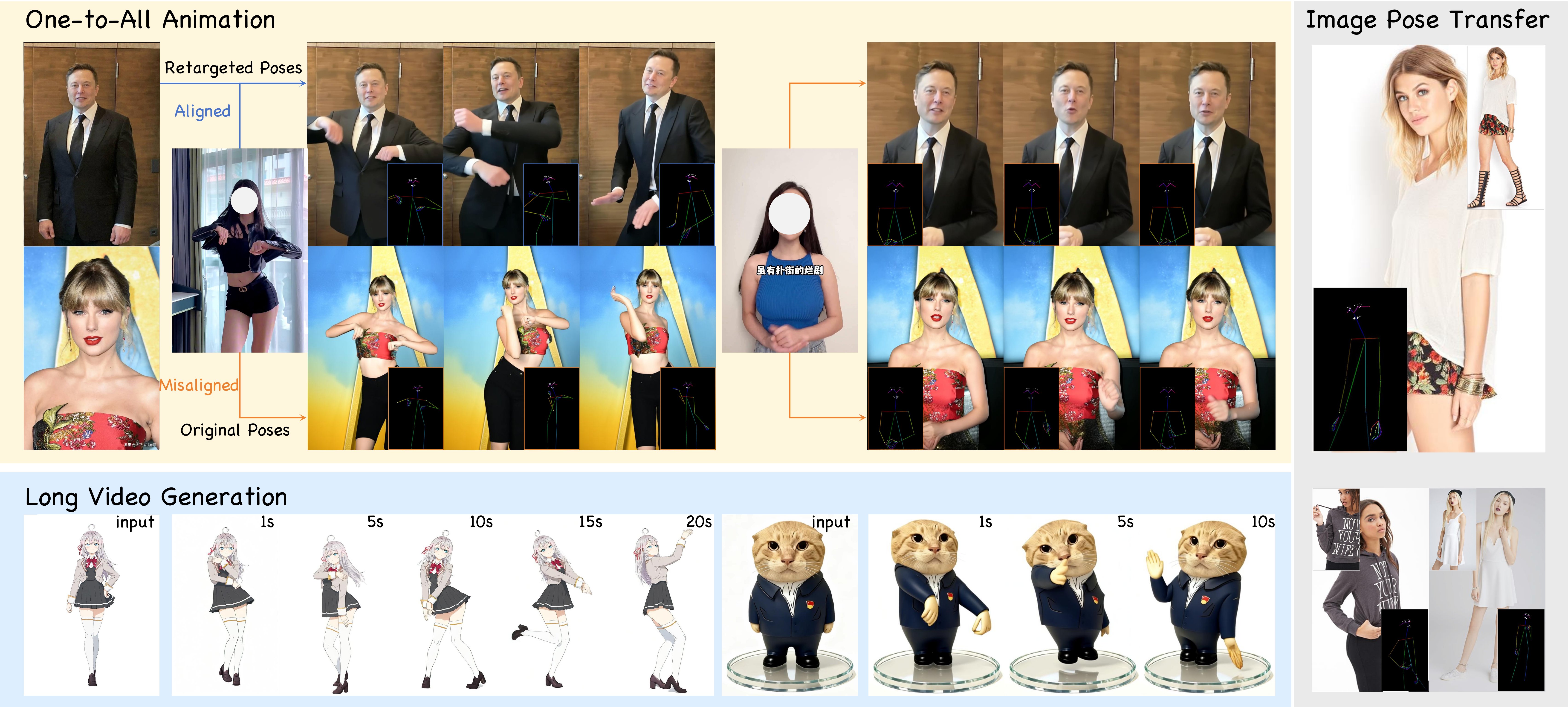}
    \vspace{-0.6cm}
    \captionof{figure}{We introduce \textit{One-to-All Animation}, a unified framework for pose-driven personalized generation. Unlike prior methods that require both spatially-aligned references and pose retargeting, our framework supports: (1) cross-scale video animation with either retargeted or original driving motion, (2) cross-scale image pose transfer, and (3) temporally coherent long video generation.}
    \label{fig:overview}
    \vspace{-0.2cm}
\end{center}
}]

\renewcommand{\thefootnote}{\fnsymbol{footnote}}
\footnotetext[0]{\hspace{-2em}$^{*}$Equal contribution: ssj180123@gmail.com, xujing0@mail.ustc.edu.cn}
\footnotetext[0]{\hspace{-2em}$^{\dagger}$Corresponding author: hukai\_wlw@jiangnan.edu.cn, 186368@zju.edu.cn}
\begin{abstract}
Recent advances in diffusion models have greatly improved pose-driven character animation.
However, existing methods are limited to spatially aligned reference-pose pairs with matched skeletal structures. Handling reference-pose misalignment remains unsolved.
To address this, we present One-to-All Animation, a unified framework for high-fidelity character animation and image pose transfer for references with arbitrary layouts.
First, to handle spatially misaligned reference, we reformulate training as a self-supervised outpainting task that transforms diverse-layout reference into a unified occluded-input format. Second, to process partially visible reference, we design a reference extractor for comprehensive identity feature extraction. Further, we integrate hybrid reference fusion attention to handle varying resolutions and dynamic sequence lengths. Finally, from the perspective of generation quality, we introduce identity-robust pose control that decouples appearance from skeletal structure to mitigate pose overfitting, and a token replace strategy for coherent long-video generation. Extensive experiments show that our method outperforms existing approaches. The code and model are available at {\small\url{https://github.com/ssj9596/One-to-All-Animation}}.
\end{abstract}   
\vspace{-0.4cm}
\section{Introduction}
In recent years, diffusion-based models have revolutionized the field of visual content generation, achieving significant improvements in both visual fidelity and controllability. These advances have opened new opportunities across film production, digital advertising, immersive virtual avatars, and other creative industries. 
Building on this progress, a growing line of research \cite{magicpose,aa, unianimate, mimicmotion,stableanimator,animatex} has focused on character animation, which generates animated videos of a given character by transferring motion from a driving video.
Recent approaches \cite{humandit, unianimatedit, wananimate} have leveraged large-scale pre-trained video foundation models built on the Diffusion Transformer (DiT) architecture \cite{dit} to achieve better temporal consistency and visual realism.

However, a critical challenge limits the practical use of character animation: the misalignment between the reference image and the driving video. 
Specifically, ``misalignment'' refers to the inconsistency of the target subject's presentation between two inputs, which manifests in two main aspects:
(1) \textbf{Spatial layout mismatch}. The reference image and driving video differ in pose scale, or body part coverage (e.g., the reference shows a half-body close-up while the driving video shows full-body dancing), and (2) \textbf{Facial inconsistency}. The reference and driving subjects differ in facial skeletal structure and geometry, particularly in the distances and proportions between facial features (eyes, nose, mouth, neck). Such misalignment often causes severe artifacts in the generated animation, including distorted body shapes and mismatched appearance.
To address these issues, an ideal solution must satisfy two requirements: spatial flexibility to handle arbitrary layout variations, and identity robustness to preserve appearance consistency despite skeletal and facial geometry differences. However, existing methods fall short in both aspects. 
Current approaches adopt a self-driven reconstruction strategy during training, inherently ensuring shared layout and skeleton between reference and driving. To maintain such alignment at inference, they impose two constraints: explicitly requiring spatial-matched reference images and strong dependence on pose retargeting to align driving poses. 

\begin{figure}
    \centering
    \includegraphics[width=1\linewidth]{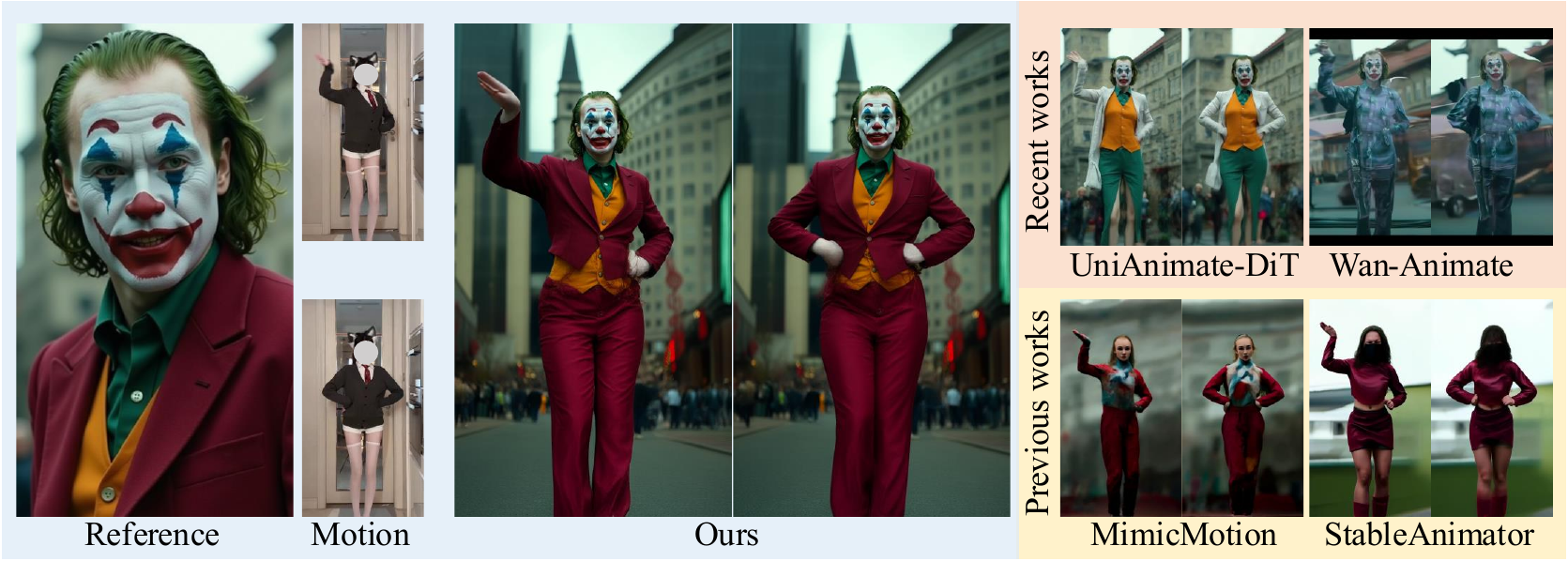}
    \vspace{-0.7cm}
    \caption{Visual comparison under spatial-misaligned inputs. Previous methods such as MimicMotion~\cite{mimicmotion} and StableAnimator~\cite{stableanimator} fail to preserve identity. Recent approaches including UniAnimate~\cite{unianimatedit} and Wan-Animate~\cite{wananimate} show degraded quality. Our method maintains robust appearance consistency.}
    \label{fig:motivation}
    \vspace{-0.7cm}
\end{figure}

As shown in Fig.~\ref{fig:motivation}, when facing mismatched inputs, previous works~\cite{mimicmotion, stableanimator} completely lose appearance information, generating results with wrong identity. Recent DiT-based methods~\cite{unianimatedit, wananimate} trained on large-scale data achieve facial preservation but still exhibit noticeable visual artifacts.
Moreover, since identity preservation relies heavily on accurate pose retargeting, its failure directly causes identity drift (See Fig.~\ref{fig:ablation_pose}). 
These limitations inspire us to rethink the training strategy. Instead of relying on perfect alignment during training, we ask: \textbf{can we train the model to handle misalignment directly?}
In this paper, we propose \emph{One-to-All Animation}, a unified framework for pose-driven personalized generation from arbitrary reference images. Our key insight is to reformulate training as an outpainting problem with a unified occluded-input format, enabling the model to learn generation from diverse spatial layouts.

Built on this insight, we introduce three key components.
First, we introduce a Reference Extractor for multi-level identity feature extraction and a hybrid reference fusion attention module to handle variable resolutions and dynamic sequence lengths. Second, we decouple identity from skeletal structure through face region enhancement and reference-guided pose control. The former randomly replaces facial structures in driving poses to break identity-skeleton coupling, while the latter balances standard and enhanced pose signals using reference appearance features. Third, we adopt a token replace strategy for long-video generation that ensures smooth cross-segment transitions.

Moreover, we extend to hybrid image-video training, naturally enabling image pose transfer as a complementary capability. 
By solving the misalignment problem, One-to-All Animation enables a single reference to adapt to multiple scenarios without alignment constraints.
As illustrated in Fig.~\ref{fig:overview}, a single arbitrary reference can drive video generation with different poses (retargeted or original), and supports cross-scale image pose transfer. Our main contributions are summarized as follows:

\begin{itemize}
    \item We present One-to-All Animation, the first unified framework for pose-driven 
    personalized image and video generation. It supports various cross-scale applications from any reference image, including character animation and image pose transfer.

    \item We design a Reference Extractor with hybrid fusion attention that enables robust identity preservation under arbitrary spatial layouts and partial visibility conditions.

    \item We introduce identity-robust pose control that decouples identity from skeletal structure to address the facial pose overfitting problem.

    \item We propose token replace training strategy for seamless long video generation.
    
\end{itemize}

\section{Related Works}
\subsection{Diffusion Models}
Diffusion models~\cite{ddim,ddpm} have demonstrated remarkable generative capabilities in both image~\cite{ldm,controlnet,controlnext,dalle2,imagen,sd3,flux} and video generation~\cite{makeavid,lvdm,svd,cogvideox,wan,ltx,hunyuanvideo,lumina,zhang2025matrix,he2025matrix,shi2025self}, transforming visual synthesis from a research topic into practical applications.
Among these, Diffusion Transformers (DiT)~\cite{dit} have become increasingly popular due to their better scalability and ability to capture long-range dependencies. On the image side, DiT-based models like SD3~\cite{sd3} and Flux~\cite{flux} achieve state-of-the-art quality in text-to-image generation. On the video side, recent foundation models~\cite{hunyuanvideo,cogvideox,wan,ltx} follow a similar pipeline: a 3D VAE~\cite{vae} first compresses video into spatio-temporal latents, which are then processed by stacked Transformer blocks for joint spatial–temporal modeling. 
Notably, many of these models are pretrained on mixed video-image datasets, which also gives them strong image generation capability. This makes them well-suited for both static image synthesis and dynamic video generation tasks, such as the pose-driven personalized generation discussed below.
\subsection{Pose-Driven Personalized Generation}
Pose-driven personalized generation aims to transfer appearance from a reference image to a specified pose. Based on the spatial alignment between two inputs, existing tasks can be divided into two settings: (\emph{i}) unaligned image-to-image generation, where the reference image and target pose differ in position, scale, and viewpoint; (\emph{ii}) aligned image-to-video generation, where the driving video poses are spatially consistent with the reference frame. 
This division stems from \textbf{data availability}: image-to-image tasks benefit from abundant training data with diverse pose variations~\cite{deepfashion}, while video generation typically relies on self-reconstruction data, naturally resulting in spatial alignment.

In the image domain, image pose transfer~\cite{poseimggan1,poseimggan2,poseimg1,poseimg2,poseimg3,poseimg4,poseimg5} corresponds to the misaligned setting, where early GAN-based methods~\cite{poseimggan1,poseimggan2} struggle with distorted textures under large pose disparities and recent diffusion-based approaches~\cite{poseimg3,poseimg4,poseimg5} remain constrained to low-resolution outputs with limited facial quality. In the video domain, character animation methods~\cite{magicpose,aa,stableanimator,mimicmotion,unianimatedit,wananimate} perform aligned image-to-video generation by animating a reference character according to a driving video. These methods ensure temporal consistency through temporal modules~\cite{animatediff} or video foundation models~\cite{svd,wan}, but still heavily rely on spatial alignment between the reference and driving poses, suffering significant degradation when this alignment is disrupted. Despite progress in both domains, existing methods are still designed for separate settings. In contrast, our method unifies unaligned image-to-image, aligned image-to-video, and unaligned image-to-video generation in a single framework.
\begin{figure*}
    \centering
    \includegraphics[width=1\linewidth]{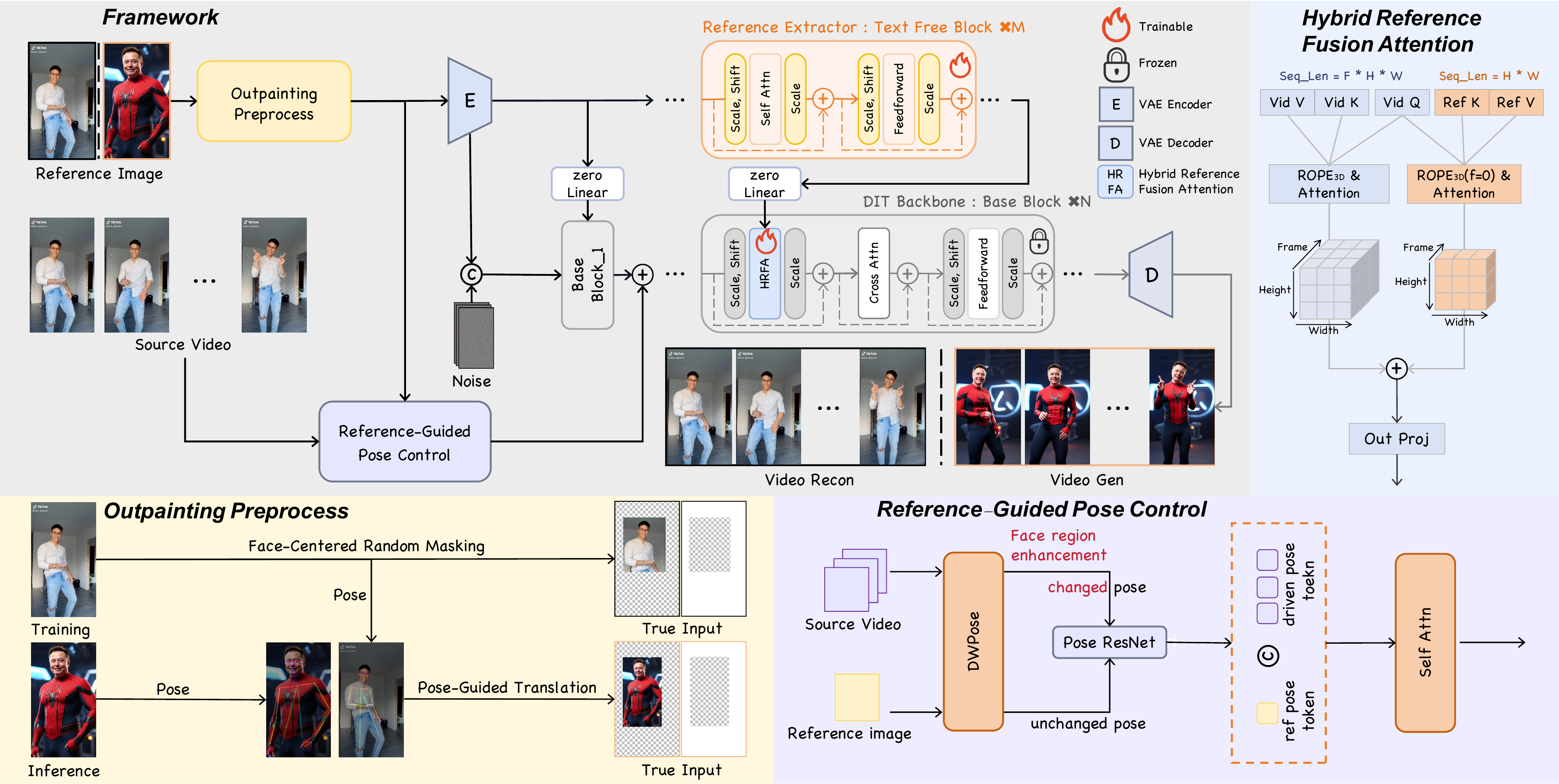}
    \vspace{-0.6cm}
    \caption{Overview of the proposed framework. We introduce outpainting preprocess to handle diverse body proportions through face-centered random masking during training and pose-guided translation at inference. The driving poses are encoded and refined via reference-guided pose control to preserve facial identity despite skeletal mismatch. Reference features are progressively injected through hybrid reference fusion attention, supporting variable resolutions and dynamic sequence lengths.}
    \label{fig:framework}
    \vspace{-0.5cm}
\end{figure*}


\section{Method}
Given a reference image and a driving video, our goal is to generate an animated video that preserves the identity from the reference while following the motion from the driving video. Unlike existing methods, which rely on assumptions such as similar camera distances and well-aligned skeletons, we explicitly tackle the challenging real-world scenarios where the reference image and driving video exhibit large scale variations, diverse spatial framings, and mismatched skeletal layouts. 
We address these challenges through both data and model design. On the data side, we introduce a self-supervised training scheme that synthesizes spatially mismatched reference-driving pairs (Sec.~\ref{sec:framework}). On the model side, we propose Reference Extractor to handle such occluded reference (Sec.~\ref{sec:ref_extractor}). Additionally, we introduce Identity-Robust Pose Control (Sec.~\ref{sec:pose_control}) to reduce pose overfitting, and TokenReplace (Sec.~\ref{sec:tokenreplace}) to enable long-video generation.
\subsection{One-to-All Animation}
\label{sec:framework}
\noindent\textbf{Inference pipeline.} An overview of the proposed \textbf{One-to-All} framework is presented in Fig.~\ref{fig:framework}.  At the inference stage, since the reference image $\mathbf{I}^r$ could have significant scale variations relative to the driving sequence $\mathbf{P}^{1:N}$, we first perform pose-guided translation between the reference image and the driving video. Specifically, we identify an anchor frame from the driving sequence with the most similar pose orientation to the reference. We estimate their scale ratio using visible body parts present in both (e.g., inter-shoulder or inter-ear distance), then resize the reference accordingly and zero-pad it to the driving video resolution, producing a spatially adjusted input $\tilde{\mathbf{I}}^r$ and a mask $\mathbf{M}^r$ indicating the padded areas. The model takes the triplet $(\tilde{\mathbf{I}}^r, \mathbf{M}^r, \mathbf{P}^{1:N})$ as input and generates a video that follows the driving motion while synthesizing the padded areas to complete the reference appearance.

\noindent\textbf{Self-supervised training via outpainting.}
We follow the self-reconstruction training setup, but with a key modification: instead of treating all frames as naturally aligned, we simulate spatial mismatches through outpainting preprocessing.
During training, we apply face-centered random masking to the reference image to simulate various scale conditions. We generate a binary outpainting mask that indicates the missing regions. The triplet of masked reference, mask and pose sequence is fed as input, with the original complete video as supervision. This training setup produces inputs identical to those generated by pose-guided translation at inference, forcing the network to learn to hallucinate occluded regions and generate coherent motions.

\subsection{Reference Extractor.} 
\label{sec:ref_extractor}
\noindent\textbf{Motivation.} The outpainting-based training presents a unique challenge: extracting reliable appearance features from severely occluded reference. Existing methods are not designed for this scenario. Previous works typically use CLIP encoders for semantic feature extraction or directly embed reference frames into I2V backbones. However, CLIP encoders focus on global representation and lack fine-grained identity details. I2V backbones are limited by the strict first-frame ``copy-paste"~\cite{phantom,moviegen} nature, which cannot effectively complete large occluded regions.
In contrast, we design a dedicated Reference Extractor that extracts multi-level appearance features from occluded references. 

\noindent\textbf{Architecture details.}
As shown in Fig.~\ref{fig:framework}, the extractor operates in parallel with the main denoising DiT backbone, producing features in the same latent space. Given reference $\tilde{\mathbf{I}}^r$ and mask $\mathbf{M}^r$, both are first encoded into latent representations $z^r$ and $z^m$ via the 3D VAE, where $\mathbf{M}^r$ is repeated along the channel dimension to match the encoder input format. These latents are concatenated along the channel dimension and converted to patch tokens:
\begin{equation}
{r}^0 = \mathrm{patchify}\left([z^r,\, z^m]_{\text{channel}}\right),
\end{equation}
forming the initial reference feature ${r}^0$. 
This feature is then refined through $M$ text-free blocks. Each block is initialized from the DiT backbone but excluding the text cross-attention.
The $M$ block outputs, together with the initial ${r}^0$, form $M\!+\!1$ reference features. These are injected into the $N$-block denoising backbone ($M < N$) through zero-initialized linear projections. Each reference feature is shared by $n$ consecutive denoising blocks, where $n = \frac{N}{M+1}$. 

\noindent\textbf{Hybrid Reference Fusion Attention.}
The key design of our Reference Extractor is the Hybrid Reference Fusion Attention (HRFA), which enables robust identity preservation across variable resolutions and dynamic sequence lengths.
Specifically, we add a new cross-attention layer on the self-attention layer in the DiT block. Given video latent $h \in \mathbb{R}^{F \times H \times W \times C}$, 
the self-attention with 3D Rotary Position Encoding (RoPE) is formulated as:
\vspace{-0.2cm}%
\begin{equation}
\small
\mathrm{Attention}(\mathrm{Q}, \mathrm{K}, \mathrm{V}) = \mathrm{softmax}\left(\frac{\mathrm{Q} (\mathrm{K})^\top}{\sqrt{d}}\right) \mathrm{V},
\end{equation}
\vspace{-0.7cm}%
\begin{equation}
\small
\mathrm{Q}=\mathrm{RoPE}_{3D}(hW_q),\quad
\mathrm{K}=\mathrm{RoPE}_{3D}(hW_k),\quad
\mathrm{V}=hW_v.
\end{equation}
where $W_q, W_k, W_v$ are learnable projection matrices.
For reference feature $r \in \mathbb{R}^{1 \times H \times W \times C}$, the new cross-attention layer is computed as:
\vspace{-0.2cm}%
\begin{equation}
\small
\mathrm{Attention}(\mathrm{Q}', \mathrm{K}', \mathrm{V}') = \mathrm{softmax}\left(\frac{\mathrm{Q}' (\mathrm{K}')^\top}{\sqrt{d}}\right) \mathrm{V}',
\end{equation}%
\vspace{-0.7cm}%
\begin{equation}
\small
\mathrm{Q}'=\mathrm{RoPE}_{3D,f=0}(hW_q),\quad
\mathrm{K}'=\mathrm{RoPE}_{3D,f=0}(rW_k'),\quad
\mathrm{V}'=rW_v'.
\end{equation}

where $W_k'$ and $W_v'$ are two newly introduced projection matrices for the reference features.
Notably, although $h$ is a video latent with frame dimension $F$, we apply $\mathrm{RoPE}_{3D}$ with $f=0$ to $h W_q$ when computing $\mathrm{Q}'$.
This design prevents the cross-attention from learning absolute frame position dependencies between the reference and video frames. As a result, the model preserves its temporal extrapolation capability and can be seamlessly applied to both image and video generation tasks.
The fused output of the self-attention and cross-attention layer is given by:
\vspace{-0.2cm}%
\begin{equation}
\small
\mathbf{z}'_{\text{fusion}} =
\mathrm{Attention}(\mathrm{Q}, \mathrm{K}, \mathrm{V}) +
\mathrm{Attention}(\mathrm{Q}', \mathrm{K}', \mathrm{V}').
\end{equation}

\subsection{Identity-Robust Pose Control} 
\label{sec:pose_control}
\noindent\textbf{Motivation.} Following most mainstream approaches, we represent the driving sequence using 2D pose keypoints and inject them into the denoising backbone as structural control. Although our outpainting-based training resolves the spatial layout mismatch at the body level, face-level train-test inconsistency still remains in the misaligned setting: during training, the reference frame and driving pose are sampled from the same video and are therefore facially aligned, whereas at inference the driving facial geometry may differ from that of the reference image. As a result, the model tends to overfit to the driving facial skeleton.

\begin{figure}
    \centering
    \includegraphics[width=1\linewidth]{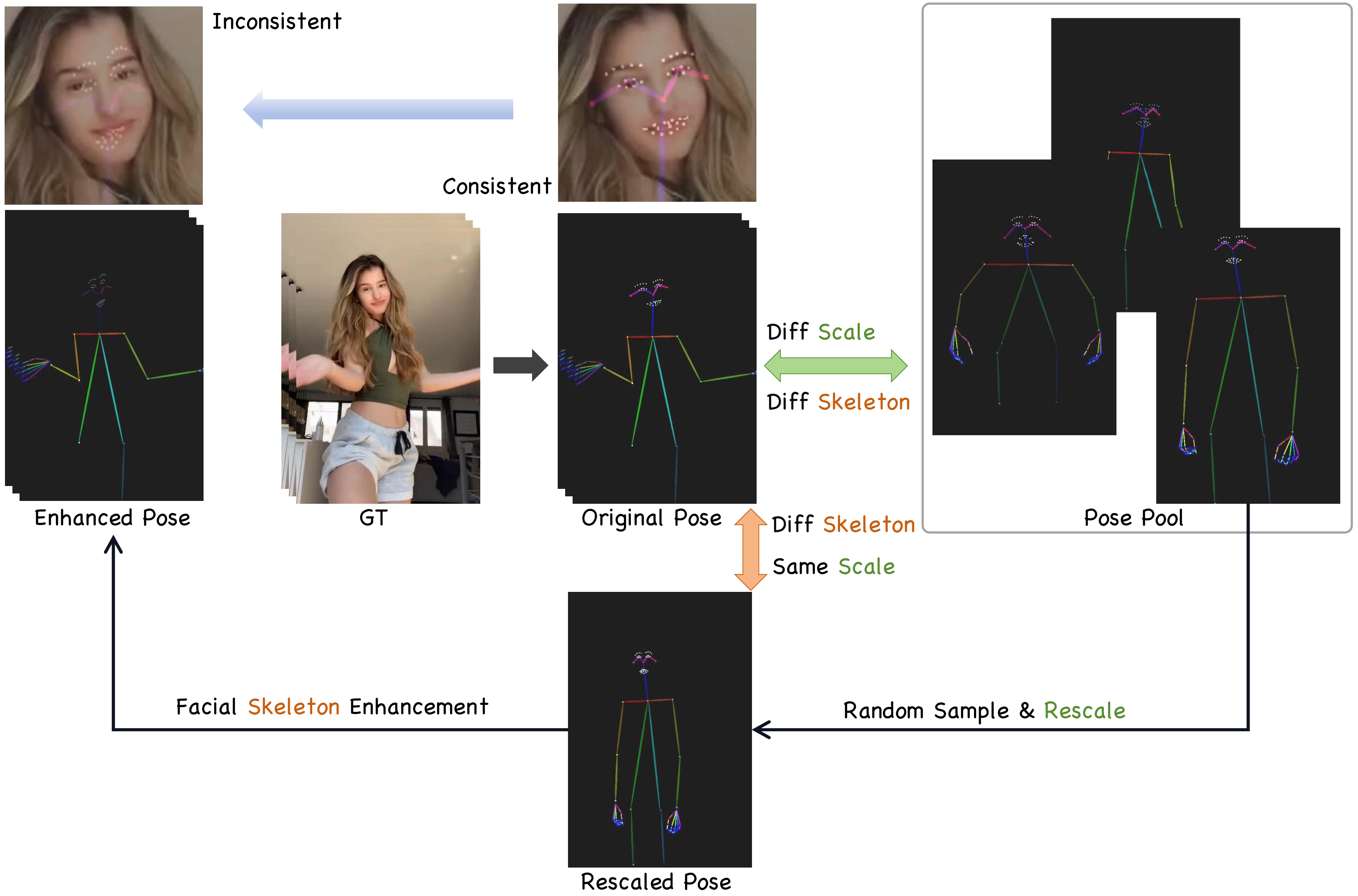}
    \vspace{-0.7cm}
    \caption{Face region enhancement. We sample a random pose, rescale it to match the driving pose, and apply their skeletal difference to create intentional facial skeleton inconsistency. A color-lighten signal indicates augmented poses.}
    \label{fig:pose}
    \vspace{-0.6cm}
\end{figure}
\noindent\textbf{Face region enhancement.} 
As illustrated in Fig.~\ref{fig:pose}, we perturb only the facial keypoints of the driving poses during training, while keeping the body keypoints unchanged. This breaks the facial alignment between the driving pose and the target frame, forcing the model to recover identity from the reference image rather than overfitting to the driving facial skeleton. We apply this enhancement to 70\% of the training samples and use a lightened-color signal to indicate whether a pose is original or enhanced.
At inference time, we retain the landmark's facial control. When retargeting is unreliable, the model can accept lightened-color pose signals to preserve identity without strict skeletal matching.

\noindent\textbf{Reference-guided pose control.}
While face region enhancement improves robustness to skeletal misalignment, it also disrupts the original pose injection (element-wisely added at each spatial location) and introduces training instability. To address this, we propose reference-guided pose control that leverages the reference image to refine the driving sequence.
Specifically, the reference image $\tilde{\mathbf{I}^r}$ is encoded by VAE to obtain its latent feature $\mathbf{z}^r$, which is concatenated with the video latents $\mathbf{z}^{1:n}$ along frame dimension:
\begin{equation}
\tilde{\mathbf{z}}^{1:(n+1)} = [\,\mathbf{z}^r,\, \mathbf{z}^{1:n}\,]_\text{frame},
\end{equation}
and fed into the DiT backbone.
In parallel, we extract the pose $\tilde{\mathbf{P}^r}$ of the reference image $\tilde{\mathbf{I}^r}$. Importantly, reference pose does not undergo enhancement and remains aligned with the reference image.
A Pose ResNet encodes both reference pose $\tilde{\mathbf{P}^r}$ and driving pose sequence $\mathbf{P}^{1:n}$ into features $\mathbf{p}^r$ and $\mathbf{p}^{1:n}$, which are also concatenated along frame dimension and processed by a self-attention (SA) block to capture intra-sequence dependencies:
\begin{equation}
\small
\tilde{\mathbf{p}}^{1:(n+1)} = \mathrm{SA}([\,\mathbf{p}^r,\, \mathbf{p}^{1:n}\,]_\text{frame}).
\end{equation}
This process propagates consistent appearance cues across frames.
The refined pose representation $\tilde{\mathbf{p}}^{1:(n+1)}$ is then element-wise added to the output of the first DiT block.

\begin{figure}
    \centering
    \includegraphics[width=1\linewidth]{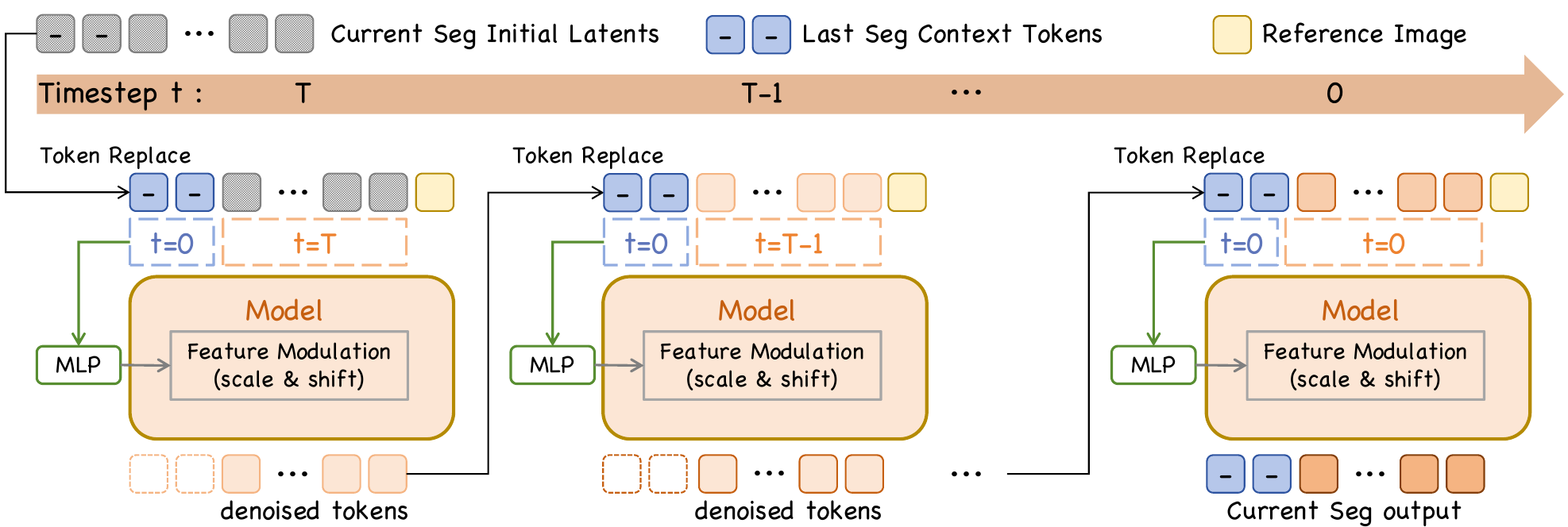}
    \vspace{-0.6cm}
    \caption{At each denoising timestep, context tokens from the last segment (\textcolor{blue}{blue}) are replaced into the current segment's initial latents, modulated with $t=0$ to serve as clean signals. This process ensures smooth transitions across segment boundaries.}
    \label{fig:tokenreplace}
    \vspace{-0.5cm}
\end{figure}

\subsection{Token Replace for Long-Video Generation.}
\label{sec:tokenreplace}
For exceedingly long video generation, the pose sequence is divided into multiple segments and generated sequentially. 
To ensure smooth transitions across segment boundaries, we adopt a token replace strategy. 
As illustrated in Fig.~\ref{fig:tokenreplace}, the last five frames of the preceding segment are encoded by VAE into two latent frames, serving as context tokens $\mathbf{z}_{\mathrm{ctx}}$. 
At each denoising timestep $t$, the first two latent frames in the noisy latent $\mathbf{z}^{1:n}_t$ are replaced by these context tokens:
\begin{equation}
\tilde{\mathbf{z}}^{1:n}_t = [\,\mathbf{z}_{\mathrm{ctx}},\, \mathbf{z}^{3:n}_t\,].
\end{equation}
During training, context tokens are excluded from the reconstruction loss and act solely as temporal guidance; at inference, they are treated as clean signals with $t=0$ in the feature modulation module throughout the denoising process. 
After denoising, the first two tokens are retained as context, and the VAE decodes the complete latent sequence to obtain the final video frames.

\begin{table*}[t]
\vspace{-0.6cm}
\caption{
Quantitative comparisons on TikTok and Cartoon datasets. In the table, $a$ / $b$ denotes results on TikTok / Cartoon. 
}
\vspace{-0.3cm}
\label{tab:video}
\centering
\setlength{\tabcolsep}{6pt}
\renewcommand{\arraystretch}{1.2}
\begin{tabular}{l | c c c c | c c}
\toprule
\textbf{Model} & \textbf{PSNR$\uparrow$} & \textbf{SSIM$\uparrow$} & \textbf{LPIPS$\downarrow$} & \textbf{FID$\downarrow$}
& \textbf{FID-VID$\downarrow$} & \textbf{FVD$\downarrow$} \\
\midrule
\rowcolor{gray!15}
\multicolumn{7}{l}{\textit{\textbf{\small $\sim$1.3B Models}}} \\
MimicMotion~\cite{mimicmotion}  & 15.43/15.09 & 0.721 / 0.647 & 0.315 / 0.368 & 67.09 / 87.06 & 18.56 / 58.35 & 412.50 / 943.38 \\
StableAnimator~\cite{stableanimator} &  14.92/15.16 & 0.737 / 0.638 & 0.315 / 0.333 & 76.50 / 70.92 & 21.57 / 31.30 & 477.28 / 720.48 \\
Animate-X~\cite{animatex} &  15.22/15.63 & 0.741 / 0.659 & 0.329 / 0.330 & \textbf{52.96} / \textbf{59.68} & \textbf{17.81} / 33.47 & 375.56 / 723.25 \\
\textbf{One-to-All-1.3B}  &\textbf{17.75}/\textbf{16.24} & \textbf{0.788} / \textbf{0.677} & \textbf{0.269} / \textbf{0.289} & 74.96 / 67.82 & 21.37 / \textbf{29.11} & \textbf{361.85} / \textbf{549.27} \\
\midrule
\rowcolor{gray!15}
\multicolumn{7}{l}{\textit{\textbf{\small $\sim$14B Models}}} \\
UniAnimate-DiT~\cite{unianimatedit} & \textbf{19.07}/17.03 & \textbf{0.816} / 0.699 & 0.265 / 0.269 & 55.32 / 55.29 & 17.42 / 24.26   & 358.42 / 510.61 \\
Wan-Animate~\cite{wananimate}  & 17.57/16.43 & 0.763 / 0.659 & 0.306 / 0.318 & 66.98 / 58.94  & 16.79 / 27.74 & \textbf{282.86} / 485.92 \\
\textbf{One-to-All-14B} & 18.07/\textbf{17.10} & 0.812 / \textbf{0.701} & \textbf{0.254} / \textbf{0.259} & \textbf{50.49} / \textbf{50.07} & \textbf{13.93} / \textbf{15.07} & 297.94 / \textbf{403.47} \\
\bottomrule
\end{tabular}
\vspace{-0.6cm}
\end{table*}

\subsection{Training and Inference}
\noindent\textbf{Training setup.} 
We train the model following Rectified Flow (RF) formulation~\cite{flowmatching,rectifiedflow}.
In the forward process, Gaussian noise $\boldsymbol{\varepsilon} \sim \mathcal{N}(0,I)$ is added to a clean latent sample $\mathbf{x}_0$ to produce:
\begin{equation}
\small
\mathbf{x}_t = (1-t)\,\mathbf{x}_0 + t\,\boldsymbol{\varepsilon},
\label{eq:xt}
\end{equation}
where the time step $t$ is obtained by sampling an integer $\tau \in \{0,\ldots,T\}$ (with $T=1000$) and normalising it to $[0,1]$.  
The network $G_{\theta}$ is trained to predict the target velocity:
\begin{equation}
u_t = \frac{\partial\mathbf{x}_t}{\partial t} = \boldsymbol{\varepsilon} - \mathbf{x}_0,
\label{eq:ut}
\end{equation}
using the regression loss:
\begin{equation}
\mathcal{L}_{\mathrm{RF}} = \left\lVert v_t - u_t \right\rVert^{2},
\label{eq:diffusion_loss}
\end{equation}
where $v_t = G_{\theta}(\mathbf{x}_t,\,t,\,C)$ and $C$ denotes the conditioning inputs (pose sequence, reference image, and mask).

\noindent\textbf{Three-stage training.} 
\label{sec:three_stage_training}
Our training process is divided into three stages. In the first stage, we train the reference extractor and the $W_k'$ / $W_v'$ components of the HRFA solely using appearance information as condition. 
In the second stage, we introduce pose condition and train the reference-guided pose control jointly with all components of the HRFA.
In the third stage, we include the token replace mechanism. 
Throughout all stages, the text prompt is fixed as an empty string.
We perform joint image-video training with mixed resolutions (512px and 768px) and various aspect ratios. The ratio of video to image data per epoch is set to 6:1. To reduce computational cost, each video sample is limited to 29 frames.
Further details are provided in the Supplementary Material.

\noindent\textbf{Inference setting.} 
At inference, we employ the Euler method for sampling over 30 steps. We adopt cumulative classifier-free guidance~\cite{pix2pix} to strengthen both the reference appearance and pose guidance. Specifically, the denoised output at each step $t-1$ is computed as:
\begin{equation}
x_{t-1} = x^{t-1}_{\varnothing}
+ \lambda_P \left( x^{t-1}_{P} - x^{t-1}_{\varnothing} \right)
+ \lambda_I \left( x^{t-1}_{RP} - x^{t-1}_{P} \right),
\end{equation}
where $x^{t-1}_{\varnothing}$ is the unconditional prediction, $x^{t-1}_{P}$ is the pose-conditioned prediction, and $x^{t-1}_{RP}$ is the joint reference-plus-pose prediction. The guidance scales $\lambda_P$ and $\lambda_I$ are both set to 1.5 in most cases.
For long video generation, the sequence is segmented into clips of 65--81 frames and processed sequentially. Starting from the second segment, the last five frames of the preceding segment are encoded via VAE into two latent frames for token replace (Sec.~\ref{sec:tokenreplace}), thereby ensuring seamless transitions across segments.
\section{Experiments}

\subsection{Implementations}
Our model is built upon the open-source text-to-video backbone Wan2.1~\cite{wan}. We train two model variants with 1.3B and 14B parameters. Experiments are conducted
on 8 NVIDIA H20 GPUs. We collect approximately 7,000 human videos from the internet and expand the training corpus by incorporating additional samples from the TikTok~\cite{tiktok}, Champ~\cite{champ}, and UBC~\cite{ubc} datasets. To enhance cross-style generalization, we curate 200 cartoon character images and synthesize animation clips using Seedance~\cite{seedance} as a supplemental training subset. We also include the DeepFashion dataset~\cite{deepfashion}, which contains 52,712 high‑resolution fashion model images, to facilitate mixed image–video training.

\begin{figure}[t]
    \centering
    \captionsetup[subfigure]{font=small, labelfont=footnotesize}
    
    \begin{subfigure}[b]{1\linewidth}
        \centering
        \includegraphics[width=\linewidth]{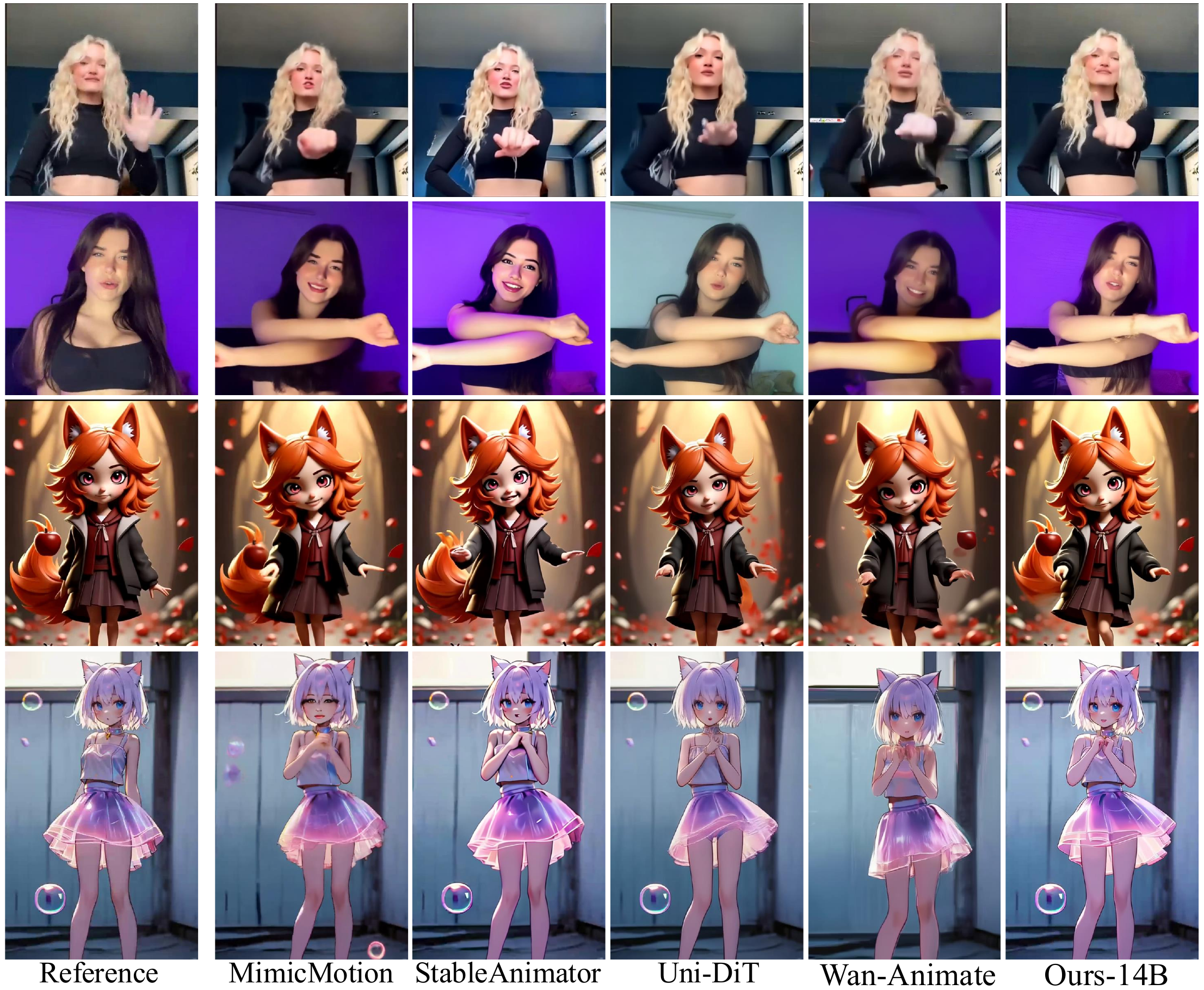}
        \caption{Results on TikTok (top two) and Cartoon testset (bottom two).}
        \label{fig:vid}
    \end{subfigure}
    
    \vspace{0.0cm}
    
    \begin{subfigure}[b]{1\linewidth}
        \centering
        \includegraphics[width=\linewidth]{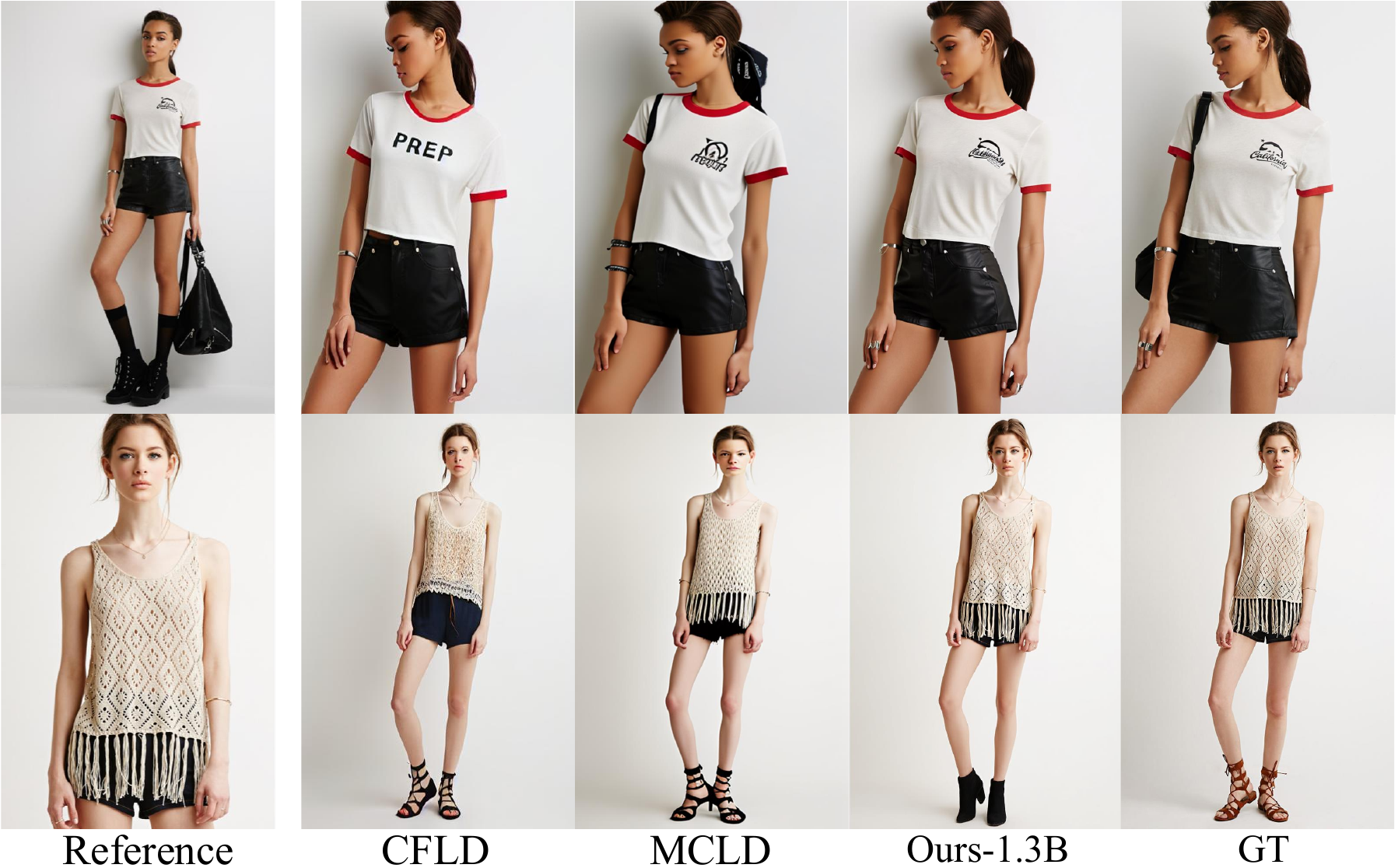}
        \caption{Results on DeepFashion benchmark.}
        \label{fig:img}
    \end{subfigure}
    
    \vspace{-0.3cm}
    \caption{Qualitative comparisons with state-of-the-art methods.}
    \label{fig:comparison}
    \vspace{-0.6cm}
\end{figure} 

\subsection{Comparisons}
\noindent\textbf{Metrics.}
We follow the previous evaluation metrics for
pose-driven personalized generation. Specifically, for single-frame quality assessment, we employ PSNR~\cite{psnr}, SSIM~\cite{ssim}, LPIPS~\cite{lpips} and FID~\cite{fid}. For video fidelity, we utilize FID-VID~\cite{fid-vid} and FVD~\cite{FVD}.

\noindent\textbf{Evaluation on Video Dataset.} For character animation, we follow previous works~\cite{disco,magicanimate} by evaluating on TikTok benchmark. In addition, we evaluate on 12 cartoon‑style image–video pairs outside the cartoon training set for cross‑domain evaluation. 
To ensure fairness, we adopt the same reference images for all methods and compare models within similar parameter scales. 
Specifically, we benchmark both $\sim$1.3B and $\sim$14B variants of our model against state-of-the-art competitors on the two datasets. 
As shown in Table~\ref{tab:video}, One-to-All-1.3B achieves superior results on most metrics among small‑scale backbones, while One-to-All-14B consistently outperforms large‑scale counterparts across both datasets, indicating strong scalability and generalization. Qualitative results are shown in Fig.~\ref{fig:vid}.

\noindent\textbf{Evaluation on Image Dataset.}
For image pose transfer, following prior works~\cite{cfld,mcld}, we evaluate on 8,570 test pairs from the DeepFashion dataset. 
Previous methods typically perform inference at a low resolution of $512\times352$, which often leads to loss of fine details. 
To explore high‑resolution generation, we additionally conduct inference at $768\,$px while maintaining the same aspect ratio ($944\times624$). 
We compare our One-to-All‑1.3B model with CFLD~\cite{cfld} and MCLD~\cite{mcld} under both resolution settings. 
As shown in Table~\ref{tab:quantresults}, our method achieves the lowest 
FID and LPIPS scores at both resolutions, indicating superior perceptual 
quality.
Although PSNR and SSIM scores are slightly lower than some 
competing methods, Fig.~\ref{fig:img} demonstrates that our approach 
produces significantly clearer facial details and better visual quality.

\begin{table}[t]
\caption{Quantitative comparison on DeepFashion dataset.}
\vspace{-0.3cm}
\label{tab:quantresults}
\centering
\small 
\setlength{\tabcolsep}{6pt}
\renewcommand{\arraystretch}{1.2}
\begin{tabular}{l | c c c c}
\toprule
\textbf{Method} & \textbf{FID$\downarrow$} & \textbf{LPIPS$\downarrow$} & 
\textbf{PSNR$\uparrow$} & \textbf{SSIM$\uparrow$} \\
\midrule
\rowcolor{gray!15}
\multicolumn{5}{l}
{\textit{\textbf{Evaluate on $512\times352$ resolution}}} \\
CFLD~\cite{cfld}  {\scriptsize\textcolor{gray}{(CVPR24)}} & 7.11 & 0.279 & \textbf{17.13} &\textbf{0.753} \\
MCLD~\cite{mcld}  {\scriptsize\textcolor{gray}{(CVPR25)}} & 7.07 & 0.275 & 16.51 & 0.736 \\
\textbf{One-to-All-1.3B} & \textbf{6.85} & \textbf{0.249} & 16.84 & 0.742 \\
\midrule
\rowcolor{gray!15}
\multicolumn{5}{l}
{\textit{\textbf{Evaluate on $944\times624$ resolution}}} \\
CFLD~\cite{cfld} {\scriptsize\textcolor{gray}{(CVPR24)}} & 8.38 & 0.314 & \textbf{17.38} & 0.758 \\
MCLD~\cite{mcld} {\scriptsize\textcolor{gray}{(CVPR25)}} & 8.96 & 0.322 & 16.33 & \textbf{0.761} \\
\textbf{One-to-All-1.3B} & \textbf{6.92} & \textbf{0.285} & 16.24& 0.754 \\
\bottomrule
\end{tabular}
\end{table}

\begin{figure}
    \centering
    \vspace{-0.3cm}
    \includegraphics[width=1\linewidth]{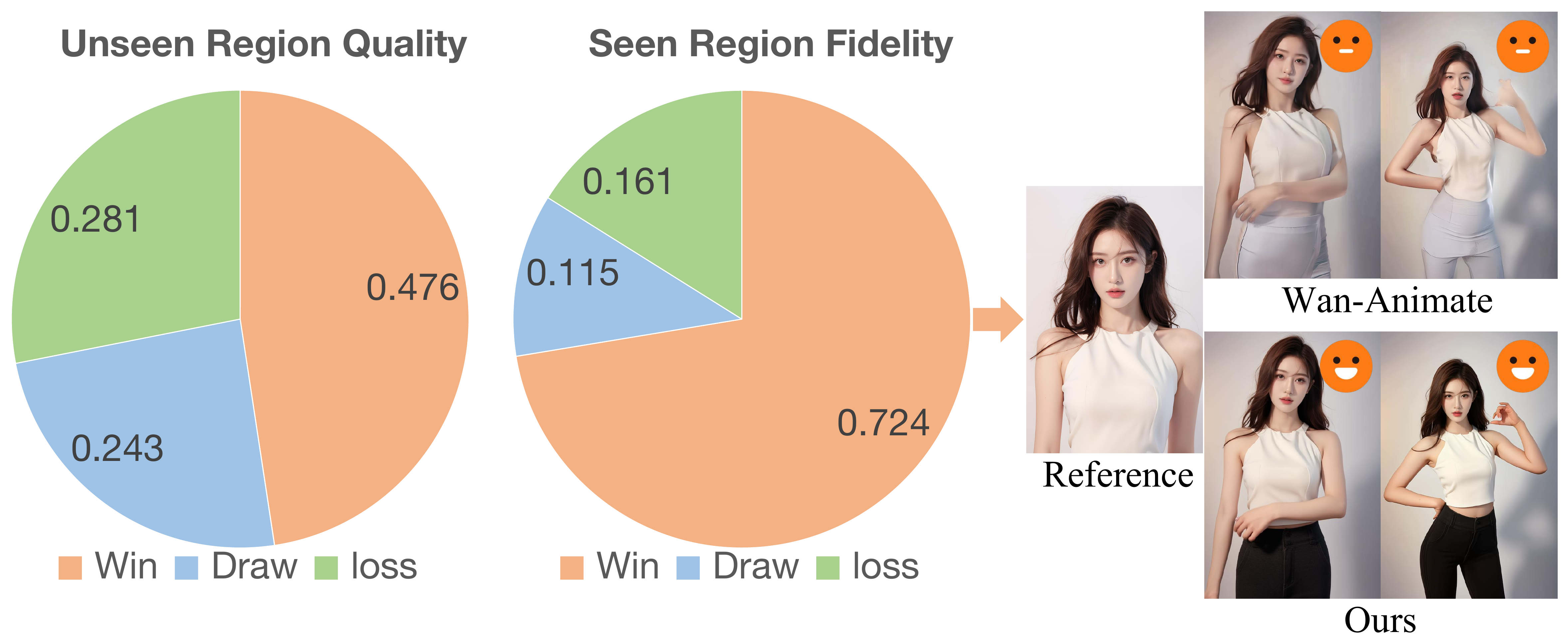}
    \vspace{-0.5cm}
    \caption{Human evaluation with current SOTA.}
    \label{fig:user_study}
    \vspace{-0.6cm}
\end{figure}

\noindent\textbf{User Study.}
To evaluate misaligned animation quality, we conduct a user study comparing with the current SOTA Wan-Animate~\cite{wananimate}. We randomly select 10 characters generated by Flux~\cite{flux} and collect 10 driving videos, resulting in 100 misaligned test cases.
Thirty participants evaluated anonymized video pairs based on: (1) Unseen region quality: visual quality and plausibility of newly generated regions; and (2) Seen region fidelity: preservation of reference appearance and background.
The results in Figure~\ref{fig:user_study} show that our method achieves superior performance (47.6\% vs 28.1\% for unseen region quality, and 72.4\% vs 16.1\% for seen region fidelity), especially in identity preservation, where it maintains consistent facial features and overall appearance better than Wan-Animate.

\subsection{Ablation Study}
\begin{figure}
    \centering
    \includegraphics[width=1\linewidth]{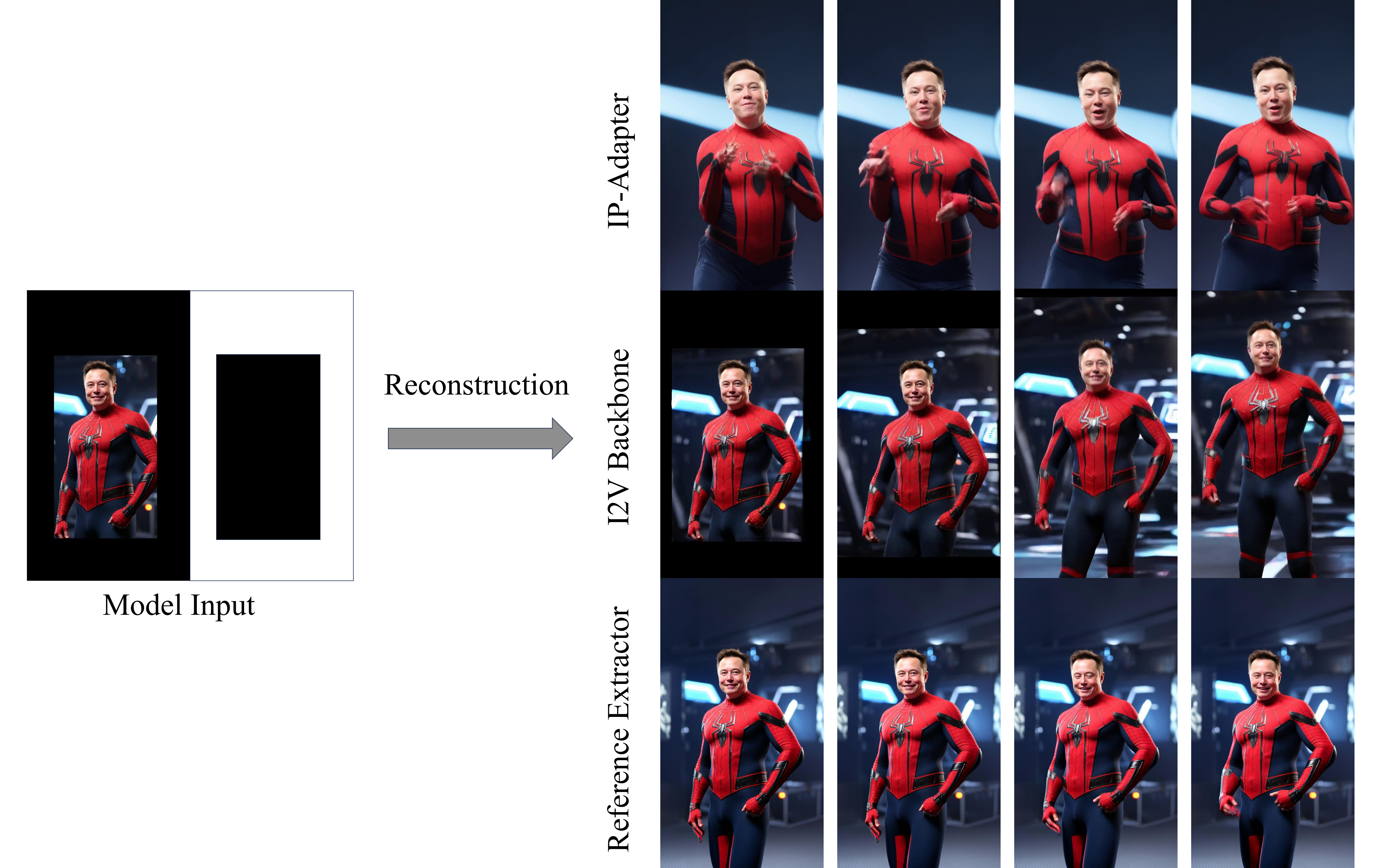}
    \vspace{-0.6cm}
    \caption{Qualitative comparison of different reference feature extraction methods in the first training stage.}
    \label{fig:ablation_ref}
    \vspace{-0.5cm}
\end{figure}

\noindent\textbf{Reference Extractor.}
To demonstrate the effectiveness of our proposed Reference Extractor, we compare against two alternatives during the first training stage: (1) IP-Adapter~\cite{ipadapter}, which uses CLIP encoder to extract semantic features, and (2) I2V 
Backbone, which relies on the model's first-frame prior for reference feature extraction.
Reconstruction results are presented in Fig.~\ref{fig:ablation_ref}. IP-Adapter exhibits poor consistency in both background and foreground details across frames. The I2V Backbone is constrained by its ``copy-paste" nature and can only gradually fill masked regions from the original input, which often leads to incomplete recovery in cases of large occlusion. In contrast, our Reference Extractor is able to reconstruct occluded regions while preserving fine-grained identity features.

\begin{figure}
    \centering
    \includegraphics[width=1\linewidth]{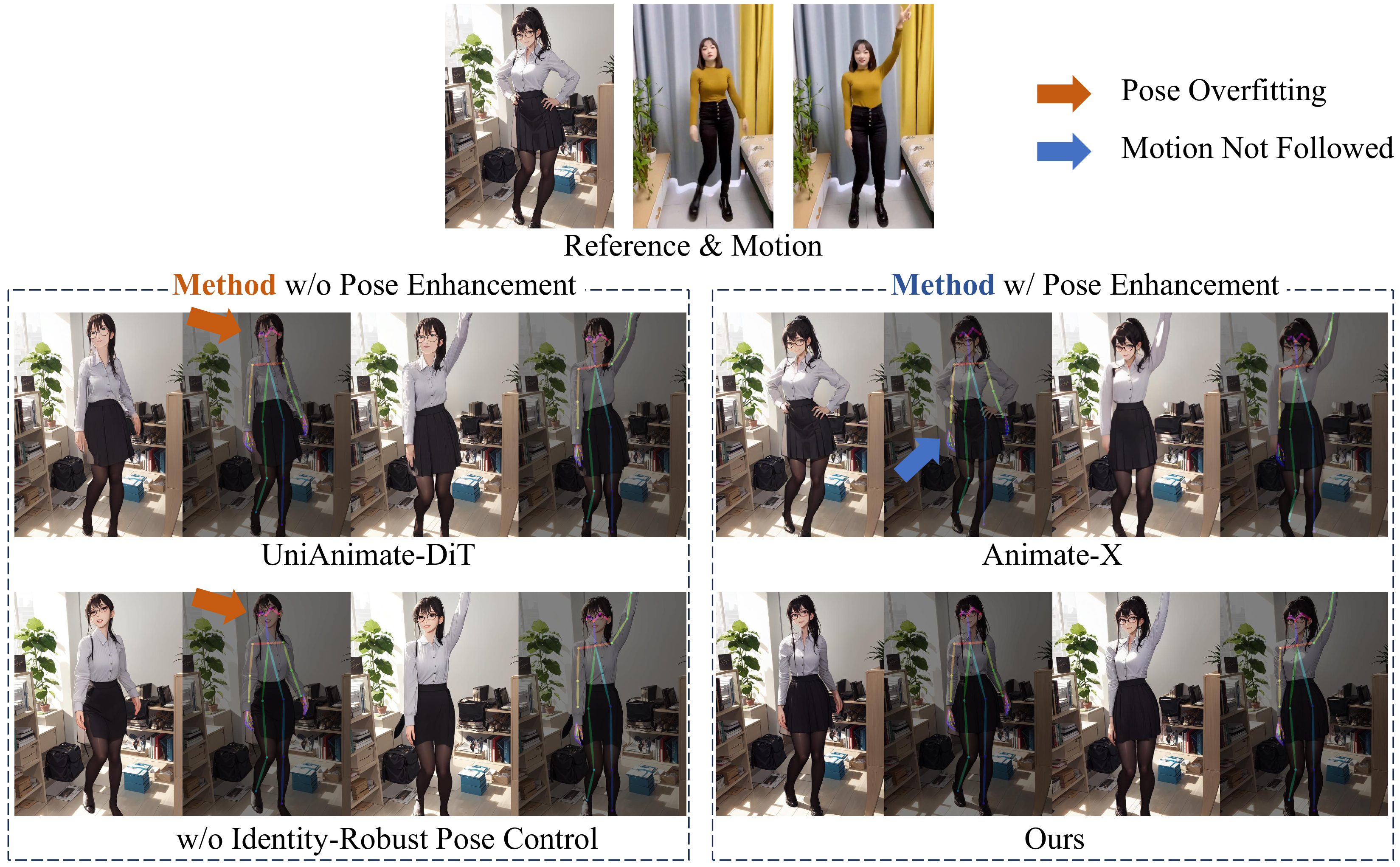}
    \vspace{-0.6cm}
    \caption{Qualitative ablation of Identity-Robust Pose Control.}
    \label{fig:ablation_pose}
    \vspace{-0.6cm}
\end{figure}

\noindent\textbf{Identity-Robust Pose Control.}
As reported in Table~\ref{tab:misaligned_ablation}, we evaluate on 100 misaligned pairs using CSIM~\cite{guo2024liveportrait}, Average Expression Distance (AED)~\cite{Siarohin_2019_NeurIPS} and Average Body Pose Distance (APD-body)~\cite{Siarohin_2019_NeurIPS}. Specifically, our full model achieves the highest identity consistency (CSIM: 0.8172). Fig.~\ref{fig:ablation_pose} provides a qualitative comparison: when pose retargeting is inaccurate, both traditional methods (e.g., UniAnimate-DiT) and our baseline without pose enhancement suffer from pose overfitting (\textcolor{orange}{orange arrows}), where the generated identity is dictated by the driving skeleton instead of the reference image.
In contrast, our method successfully preserves identity under such conditions.

\noindent\textbf{1) Face region enhancement.} While we are not the first to employ pose augmentation, our approach differs notably from prior work.
We compare against Animate-X~\cite{animatex}, which rescales the entire pose skeleton. However, such aggressive whole-body augmentation disrupts spatial correspondence across all body parts, causing motion deviation (\textcolor{blue}{blue arrows} in Fig.~\ref{fig:ablation_pose}). In contrast, ours preserves body motion accuracy while improving identity robustness.
According to Table~\ref{tab:misaligned_ablation}, this augmentation slightly increases AED compared to the baseline. However, our method still retains more expression information than approaches that completely discard facial signals (e.g., UniAnimate-DiT).
\noindent\textbf{2) Reference-guided pose control.} 
Tables~\ref{tab:ablation} and~\ref{tab:misaligned_ablation} reveal a critical insight: adding face region enhancement alone actually degrades performance. The reason is straightforward: while we intentionally introduce skeletal inconsistencies to prevent pose overfitting, the driving pose is still directly added to the noisy latent. This mismatch between enhanced pose and ground-truth causes training instability. 
Our reference-guided pose control addresses this through relationship modeling.
By concatenating the unchanged reference pose with the driving sequence, the model learns to capture their structural dependencies and adaptively refine the poses accordingly.
This stabilizes training and yields substantial improvements.

\begin{table}[t]
\centering
\caption{Ablation study on model components. Experiments are conducted on the TikTok benchmark using 14B model.}
\vspace{-0.4cm}
\resizebox{\linewidth}{!}{%
\begin{tabular}{l c c c}
\toprule
\textbf{Components} & \textbf{SSIM$\uparrow$} & \textbf{LPIPS$\downarrow$} & \textbf{FVD$\downarrow$} \\
\midrule
\makebox[\widthof{\phantom{Full  +}}][l]{Base  } Ref. Extractor & 0.773 & 0.280 & 355.2 \\
\phantom{Full  }+ Face region enhancement & 0.748 & 0.335 & 412.8 \\
\phantom{Full  }+ Reference-guided pose control & 0.795 & 0.275 & 325.7 \\
Full  + Token Replace & \textbf{0.812} & \textbf{0.254} & \textbf{297.9} \\
\bottomrule
\end{tabular}}
\label{tab:ablation}
\vspace{-0.4cm}
\end{table}

\begin{table}[t]
\centering
\caption{Ablation on identity-robust pose control using 100 misaligned image-video pairs.}
\label{tab:misaligned_ablation}
\small
\vspace{-0.4cm}
\resizebox{\linewidth}{!}{%
\begin{tabular}{lccc}
\toprule
\textbf{Method} & \textbf{CSIM}$\uparrow$ & \textbf{APD-body}$\downarrow$ & \textbf{AED}$\downarrow$ \\
\midrule
w/o identity-robust pose control & 0.6761 & 0.0358 & 0.6898 \\
w/ face region enhancement only & 0.7569 & 0.0772 & 0.9274 \\
w/ full (face enhancement + ref-guided control) & 0.8172 & 0.0367 & 0.7457 \\
UniAnimate-DiT & -- & -- & 0.8924 \\
\bottomrule
\end{tabular}}
\vspace{-0.4cm}
\end{table}

\begin{figure}
    \centering
    \includegraphics[width=1\linewidth]{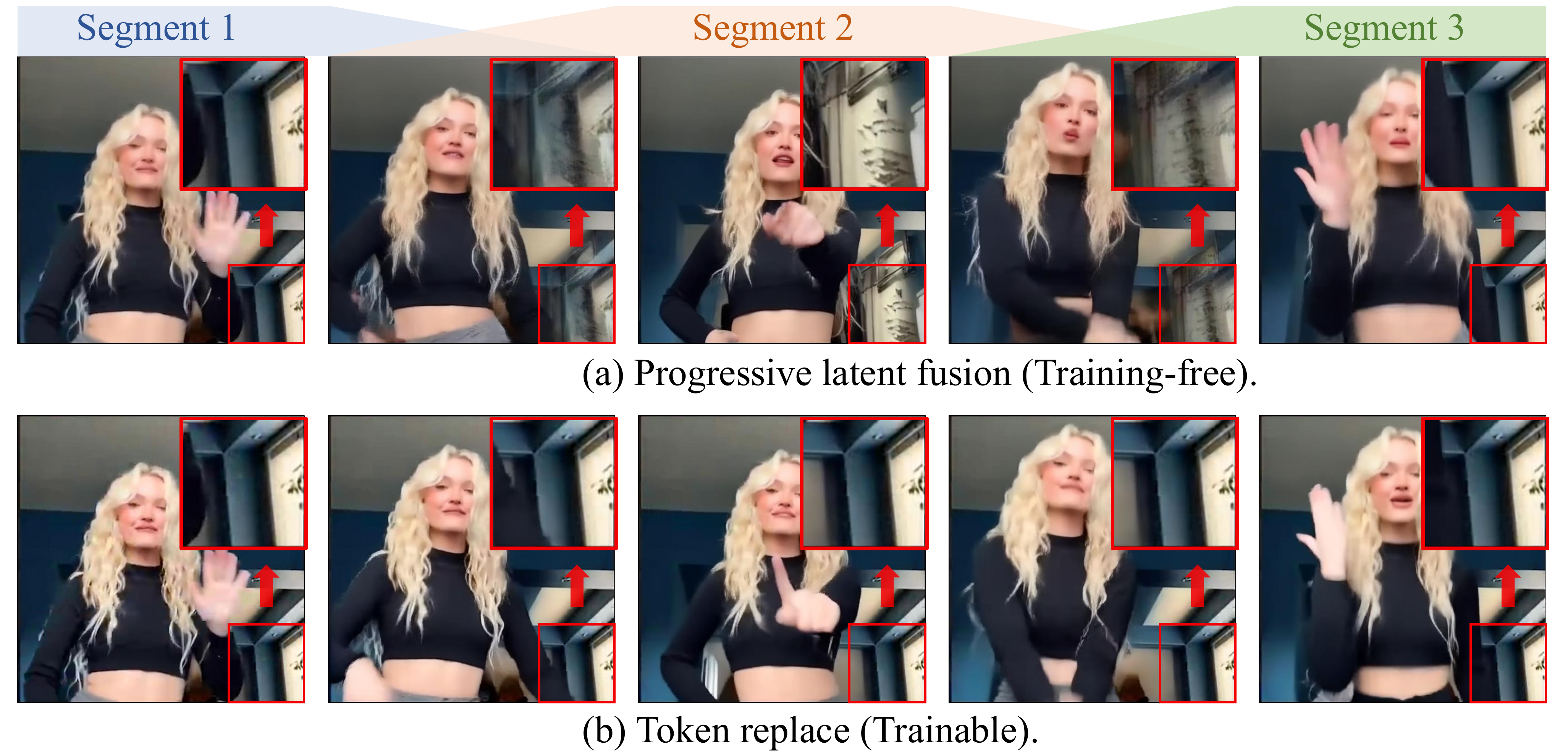}
    \vspace{-0.8cm}
    \caption{Comparison of long video generation strategies.}
    \label{fig:tokenreplace_abl}
    \vspace{-0.6cm}
\end{figure}

\noindent\textbf{Token Replace.}
We conduct a comparison with the default long video generation strategy: progressive latent fusion. This training-free approach generates video segments separately and fuses their overlapping frames at each denoising step~\cite{mimicmotion}. However, as shown in Fig.~\hyperref[fig:tokenreplace_abl]{\ref{fig:tokenreplace_abl}a}, the overlap regions fail to maintain consistency with both adjacent segments. This exposes the inherent limitation of training-free fusion approaches. In contrast, our trainable token replace strategy leverages frames from overlap regions as context token, ensuring strong temporal consistency across segments. Quantitative results in Table~\ref{tab:ablation} demonstrate that incorporating token replace achieves the best performance across all metrics.

\section{Conclusion}
In this paper, we present \emph{One-to-All Animation}, a unified framework for pose-driven personalized generation under diverse misalignment scenarios. We cast training as a self-supervised outpainting task, enabling the model to handle arbitrary spatial layouts. We design a Reference Extractor with hybrid fusion attention to preserve fine-grained identity features across variable resolutions and frames. We propose Identity-Robust Pose Control to decouple facial appearance from skeletal structure and prevent pose overfitting. Moreover, we introduce Token Replace to ensure temporal consistency in long video generation. Extensive experiments and ablation studies validate our method's robustness across diverse layouts and identity preservation, demonstrating practical flexibility for real-world applications.
{
    \small
    \bibliographystyle{ieeenat_fullname}
    \bibliography{main}

@String(CVPR= {IEEE Conf. Comput. Vis. Pattern Recog.})

@String(ICLR = {Int. Conf. Learn. Represent.})

@String(IJCAI = {IJCAI})

@String(CVPR  = {CVPR})

@String(ICLR  = {ICLR})

@article{zhang2025matrix,
  title={Matrix-Game: Interactive World Foundation Model},
  author={Zhang, Yifan and Peng, Chunli and Wang, Boyang and Wang, Puyi and Zhu, Qingcheng and Kang, Fei and Jiang, Biao and Gao, Zedong and Li, Eric and Liu, Yang and others},
  journal={arXiv preprint arXiv:2506.18701},
  year={2025}
}

@article{he2025matrix,
  title={Matrix-game 2.0: An open-source, real-time, and streaming interactive world model},
  author={He, Xianglong and Peng, Chunli and Liu, Zexiang and Wang, Boyang and Zhang, Yifan and Cui, Qi and Kang, Fei and Jiang, Biao and An, Mengyin and Ren, Yangyang and others},
  journal={arXiv preprint arXiv:2508.13009},
  year={2025}
}

@inproceedings{aa,
  title={Animate anyone: Consistent and controllable image-to-video synthesis for character animation},
  author={Hu, Li},
  booktitle={Proceedings of the IEEE/CVF Conference on Computer Vision and Pattern Recognition},
  pages={8153--8163},
  year={2024}
}

@article{unianimate,
  title={Unianimate: Taming unified video diffusion models for consistent human image animation},
  author={Wang, Xiang and Zhang, Shiwei and Gao, Changxin and Wang, Jiayu and Zhou, Xiaoqiang and Zhang, Yingya and Yan, Luxin and Sang, Nong},
  journal={arXiv preprint arXiv:2406.01188},
  year={2024}
}

@article{mimicmotion,
  title={Mimicmotion: High-quality human motion video generation with confidence-aware pose guidance},
  author={Zhang, Yuang and Gu, Jiaxi and Wang, Li-Wen and Wang, Han and Cheng, Junqi and Zhu, Yuefeng and Zou, Fangyuan},
  journal={arXiv preprint arXiv:2406.19680},
  year={2024}
}

@article{unianimatedit,
  title={Unianimate-dit: Human image animation with large-scale video diffusion transformer},
  author={Wang, Xiang and Zhang, Shiwei and Tang, Longxiang and Zhang, Yingya and Gao, Changxin and Wang, Yuehuan and Sang, Nong},
  journal={arXiv preprint arXiv:2504.11289},
  year={2025}
}

@inproceedings{stableanimator,
  title={Stableanimator: High-quality identity-preserving human image animation},
  author={Tu, Shuyuan and Xing, Zhen and Han, Xintong and Cheng, Zhi-Qi and Dai, Qi and Luo, Chong and Wu, Zuxuan},
  booktitle={Proceedings of the Computer Vision and Pattern Recognition Conference},
  pages={21096--21106},
  year={2025}
}

@inproceedings{dit,
  title={Scalable diffusion models with transformers},
  author={Peebles, William and Xie, Saining},
  booktitle={Proceedings of the IEEE/CVF international conference on computer vision},
  pages={4195--4205},
  year={2023}
}

@article{cogvideox,
  title={Cogvideox: Text-to-video diffusion models with an expert transformer},
  author={Yang, Zhuoyi and Teng, Jiayan and Zheng, Wendi and Ding, Ming and Huang, Shiyu and Xu, Jiazheng and Yang, Yuanming and Hong, Wenyi and Zhang, Xiaohan and Feng, Guanyu and others},
  journal={arXiv preprint arXiv:2408.06072},
  year={2024}
}

@article{hunyuanvideo,
  title={Hunyuanvideo: A systematic framework for large video generative models},
  author={Kong, Weijie and Tian, Qi and Zhang, Zijian and Min, Rox and Dai, Zuozhuo and Zhou, Jin and Xiong, Jiangfeng and Li, Xin and Wu, Bo and Zhang, Jianwei and others},
  journal={arXiv preprint arXiv:2412.03603},
  year={2024}
}

@article{ltx,
  title={Ltx-video: Realtime video latent diffusion},
  author={HaCohen, Yoav and Chiprut, Nisan and Brazowski, Benny and Shalem, Daniel and Moshe, Dudu and Richardson, Eitan and Levin, Eran and Shiran, Guy and Zabari, Nir and Gordon, Ori and others},
  journal={arXiv preprint arXiv:2501.00103},
  year={2024}
}

@article{wan,
  title={Wan: Open and advanced large-scale video generative models},
  author={Wan, Team and Wang, Ang and Ai, Baole and Wen, Bin and Mao, Chaojie and Xie, Chen-Wei and Chen, Di and Yu, Feiwu and Zhao, Haiming and Yang, Jianxiao and others},
  journal={arXiv preprint arXiv:2503.20314},
  year={2025}
}

@article{humandit,
  title={Humandit: Pose-guided diffusion transformer for long-form human motion video generation},
  author={Gan, Qijun and Ren, Yi and Zhang, Chen and Ye, Zhenhui and Xie, Pan and Yin, Xiang and Yuan, Zehuan and Peng, Bingyue and Zhu, Jianke},
  journal={arXiv preprint arXiv:2502.04847},
  year={2025}
}

@article{wananimate,
  title={Wan-Animate: Unified Character Animation and Replacement with Holistic Replication},
  author={Cheng, Gang and Gao, Xin and Hu, Li and Hu, Siqi and Huang, Mingyang and Ji, Chaonan and Li, Ju and Meng, Dechao and Qi, Jinwei and Qiao, Penchong and others},
  journal={arXiv preprint arXiv:2509.14055},
  year={2025}
}

@inproceedings{ddim,
  title={Denoising Diffusion Implicit Models},
  author={Song, Jiaming and Meng, Chenlin and Ermon, Stefano},
  booktitle={International Conference on Learning Representations}
}

@article{ddpm,
  title={Denoising diffusion probabilistic models},
  author={Ho, Jonathan and Jain, Ajay and Abbeel, Pieter},
  journal={Advances in neural information processing systems},
  volume={33},
  pages={6840--6851},
  year={2020}
}

@inproceedings{ldm,
  title={High-resolution image synthesis with latent diffusion models},
  author={Rombach, Robin and Blattmann, Andreas and Lorenz, Dominik and Esser, Patrick and Ommer, Bj{\"o}rn},
  booktitle={Proceedings of the IEEE/CVF conference on computer vision and pattern recognition},
  pages={10684--10695},
  year={2022}
}

@inproceedings{controlnet,
  title={Adding conditional control to text-to-image diffusion models},
  author={Zhang, Lvmin and Rao, Anyi and Agrawala, Maneesh},
  booktitle={Proceedings of the IEEE/CVF international conference on computer vision},
  pages={3836--3847},
  year={2023}
}

@article{controlnext,
  title={Controlnext: Powerful and efficient control for image and video generation},
  author={Peng, Bohao and Wang, Jian and Zhang, Yuechen and Li, Wenbo and Yang, Ming-Chang and Jia, Jiaya},
  journal={arXiv preprint arXiv:2408.06070},
  year={2024}
}

@article{dalle2,
  title={Hierarchical text-conditional image generation with clip latents},
  author={Ramesh, Aditya and Dhariwal, Prafulla and Nichol, Alex and Chu, Casey and Chen, Mark},
  journal={arXiv preprint arXiv:2204.06125},
  volume={1},
  number={2},
  pages={3},
  year={2022}
}

@article{imagen,
  title={Photorealistic text-to-image diffusion models with deep language understanding},
  author={Saharia, Chitwan and Chan, William and Saxena, Saurabh and Li, Lala and Whang, Jay and Denton, Emily L and Ghasemipour, Kamyar and Gontijo Lopes, Raphael and Karagol Ayan, Burcu and Salimans, Tim and others},
  journal={Advances in neural information processing systems},
  volume={35},
  pages={36479--36494},
  year={2022}
}

@article{makeavid,
  title={Make-a-video: Text-to-video generation without text-video data},
  author={Singer, Uriel and Polyak, Adam and Hayes, Thomas and Yin, Xi and An, Jie and Zhang, Songyang and Hu, Qiyuan and Yang, Harry and Ashual, Oron and Gafni, Oran and others},
  journal={arXiv preprint arXiv:2209.14792},
  year={2022}
}

@article{lvdm,
  title={Latent video diffusion models for high-fidelity long video generation},
  author={He, Yingqing and Yang, Tianyu and Zhang, Yong and Shan, Ying and Chen, Qifeng},
  journal={arXiv preprint arXiv:2211.13221},
  year={2022}
}

@inproceedings{animatediff,
  title={ANIMATEDIFF: ANIMATE YOUR PERSONALIZED TEXT-TO-IMAGE DIFFUSION MODELS WITHOUT SPECIFIC TUNING},
  author={Guo, Yuwei and Yang, Ceyuan and Rao, Anyi and Liang, Zhengyang and Wang, Yaohui and Qiao, Yu and Agrawala, Maneesh and Lin, Dahua and Dai, Bo},
  booktitle={12th International Conference on Learning Representations, ICLR 2024},
  year={2024}
}

@article{svd,
  title={Stable Video Diffusion: Scaling Latent Video Diffusion Models to Large Datasets},
  author={Blattmann, Andreas and Dockhorn, Tim and Kulal, Sumith and Mendelevitch, Daniel and Kilian, Maciej and Lorenz, Dominik and Levi, Yam and English, Zion and Voleti, Vikram and Letts, Adam and others},
  journal={CoRR},
  year={2023}
}

@inproceedings{sd3,
  title={Scaling rectified flow transformers for high-resolution image synthesis},
  author={Esser, Patrick and Kulal, Sumith and Blattmann, Andreas and Entezari, Rahim and M{\"u}ller, Jonas and Saini, Harry and Levi, Yam and Lorenz, Dominik and Sauer, Axel and Boesel, Frederic and others},
  booktitle={Forty-first international conference on machine learning},
  year={2024}
}

@misc{flux,
  author = {Black Forest Labs},
  title = {Flux},
  howpublished = {\url{https://github.com/black-forest-labs/flux}},
  year = {2024}
}

@article{lumina,
  title={Lumina-video: Efficient and flexible video generation with multi-scale next-dit},
  author={Liu, Dongyang and Li, Shicheng and Liu, Yutong and Li, Zhen and Wang, Kai and Li, Xinyue and Qin, Qi and Liu, Yufei and Xin, Yi and Li, Zhongyu and others},
  journal={arXiv preprint arXiv:2502.06782},
  year={2025}
}

@article{vae,
  title={Auto-encoding variational bayes},
  author={Kingma, Diederik P and Welling, Max},
  journal={arXiv preprint arXiv:1312.6114},
  year={2013}
}

@inproceedings{poseimg1,
  title={Exploring dual-task correlation for pose guided person image generation},
  author={Zhang, Pengze and Yang, Lingxiao and Lai, Jian-Huang and Xie, Xiaohua},
  booktitle={Proceedings of the IEEE/CVF conference on Computer Vision and Pattern Recognition},
  pages={7713--7722},
  year={2022}
}

@inproceedings{poseimg2,
  title={Neural texture extraction and distribution for controllable person image synthesis},
  author={Ren, Yurui and Fan, Xiaoqing and Li, Ge and Liu, Shan and Li, Thomas H},
  booktitle={Proceedings of the IEEE/CVF conference on computer vision and pattern recognition},
  pages={13535--13544},
  year={2022}
}

@inproceedings{poseimg3,
  title={Advancing Pose-Guided Image Synthesis with Progressive Conditional Diffusion Models},
  author={Shen, Fei and Ye, Hu and Zhang, Jun and Wang, Cong and Han, Xiao and Wei, Yang},
  booktitle={The Twelfth International Conference on Learning Representations}
}

@inproceedings{poseimg4,
  title={Coarse-to-fine latent diffusion for pose-guided person image synthesis},
  author={Lu, Yanzuo and Zhang, Manlin and Ma, Andy J and Xie, Xiaohua and Lai, Jianhuang},
  booktitle={Proceedings of the IEEE/CVF Conference on Computer Vision and Pattern Recognition},
  pages={6420--6429},
  year={2024}
}

@inproceedings{poseimg5,
  title={Multi-focal Conditioned Latent Diffusion for Person Image Synthesis},
  author={Liu, Jiaqi and Zhang, Jichao and Rota, Paolo and Sebe, Nicu},
  booktitle={Proceedings of the Computer Vision and Pattern Recognition Conference},
  pages={16019--16028},
  year={2025}
}

@inproceedings{poseimggan1,
  title={Deformable gans for pose-based human image generation},
  author={Siarohin, Aliaksandr and Sangineto, Enver and Lathuiliere, St{\'e}phane and Sebe, Nicu},
  booktitle={Proceedings of the IEEE conference on computer vision and pattern recognition},
  pages={3408--3416},
  year={2018}
}

@inproceedings{poseimggan2,
  title={Deep image spatial transformation for person image generation},
  author={Ren, Yurui and Yu, Xiaoming and Chen, Junming and Li, Thomas H and Li, Ge},
  booktitle={Proceedings of the IEEE/CVF conference on computer vision and pattern recognition},
  pages={7690--7699},
  year={2020}
}

@article{magicpose,
  title={Magicpose: Realistic human poses and facial expressions retargeting with identity-aware diffusion},
  author={Chang, Di and Shi, Yichun and Gao, Quankai and Fu, Jessica and Xu, Hongyi and Song, Guoxian and Yan, Qing and Zhu, Yizhe and Yang, Xiao and Soleymani, Mohammad},
  journal={arXiv preprint arXiv:2311.12052},
  year={2023}
}

@article{animatex,
  title={Animate-x: Universal character image animation with enhanced motion representation},
  author={Tan, Shuai and Gong, Biao and Wang, Xiang and Zhang, Shiwei and Zheng, Dandan and Zheng, Ruobing and Zheng, Kecheng and Chen, Jingdong and Yang, Ming},
  journal={arXiv preprint arXiv:2410.10306},
  year={2024}
}

@inproceedings{tiktok,
  title={Learning high fidelity depths of dressed humans by watching social media dance videos},
  author={Jafarian, Yasamin and Park, Hyun Soo},
  booktitle={Proceedings of the IEEE/CVF Conference on Computer Vision and Pattern Recognition},
  pages={12753--12762},
  year={2021}
}

@article{ubc,
  title={Dwnet: Dense warp-based network for pose-guided human video generation},
  author={Zablotskaia, Polina and Siarohin, Aliaksandr and Zhao, Bo and Sigal, Leonid},
  journal={arXiv preprint arXiv:1910.09139},
  year={2019}
}

@article{seedance,
  title={Seedance 1.0: Exploring the Boundaries of Video Generation Models},
  author={Gao, Yu and Guo, Haoyuan and Hoang, Tuyen and Huang, Weilin and Jiang, Lu and Kong, Fangyuan and Li, Huixia and Li, Jiashi and Li, Liang and Li, Xiaojie and others},
  journal={arXiv preprint arXiv:2506.09113},
  year={2025}
}

@inproceedings{champ,
  title={Champ: Controllable and consistent human image animation with 3d parametric guidance},
  author={Zhu, Shenhao and Chen, Junming Leo and Dai, Zuozhuo and Dong, Zilong and Xu, Yinghui and Cao, Xun and Yao, Yao and Zhu, Hao and Zhu, Siyu},
  booktitle={European Conference on Computer Vision},
  pages={145--162},
  year={2024},
  organization={Springer}
}

@inproceedings{deepfashion,
  title={Deepfashion: Powering robust clothes recognition and retrieval with rich annotations},
  author={Liu, Ziwei and Luo, Ping and Qiu, Shi and Wang, Xiaogang and Tang, Xiaoou},
  booktitle={Proceedings of the IEEE conference on computer vision and pattern recognition},
  pages={1096--1104},
  year={2016}
}

@inproceedings{disco,
  title={Disco: Disentangled control for realistic human dance generation},
  author={Wang, Tan and Li, Linjie and Lin, Kevin and Zhai, Yuanhao and Lin, Chung-Ching and Yang, Zhengyuan and Zhang, Hanwang and Liu, Zicheng and Wang, Lijuan},
  booktitle={Proceedings of the IEEE/CVF Conference on Computer Vision and Pattern Recognition},
  pages={9326--9336},
  year={2024}
}

@inproceedings{magicanimate,
  title={Magicanimate: Temporally consistent human image animation using diffusion model},
  author={Xu, Zhongcong and Zhang, Jianfeng and Liew, Jun Hao and Yan, Hanshu and Liu, Jia-Wei and Zhang, Chenxu and Feng, Jiashi and Shou, Mike Zheng},
  booktitle={Proceedings of the IEEE/CVF Conference on Computer Vision and Pattern Recognition},
  pages={1481--1490},
  year={2024}
}

@inproceedings{cfld,
  title={Coarse-to-fine latent diffusion for pose-guided person image synthesis},
  author={Lu, Yanzuo and Zhang, Manlin and Ma, Andy J and Xie, Xiaohua and Lai, Jianhuang},
  booktitle={Proceedings of the IEEE/CVF Conference on Computer Vision and Pattern Recognition},
  pages={6420--6429},
  year={2024}
}

@article{flowmatching,
  title={Flow matching for generative modeling},
  author={Lipman, Yaron and Chen, Ricky TQ and Ben-Hamu, Heli and Nickel, Maximilian and Le, Matt},
  journal={arXiv preprint arXiv:2210.02747},
  year={2022}
}

@article{rectifiedflow,
  title={Flow straight and fast: Learning to generate and transfer data with rectified flow},
  author={Liu, Xingchao and Gong, Chengyue and Liu, Qiang},
  journal={arXiv preprint arXiv:2209.03003},
  year={2022}
}

@inproceedings{pix2pix,
  title={Instructpix2pix: Learning to follow image editing instructions},
  author={Brooks, Tim and Holynski, Aleksander and Efros, Alexei A},
  booktitle={Proceedings of the IEEE/CVF conference on computer vision and pattern recognition},
  pages={18392--18402},
  year={2023}
}

@inproceedings{psnr,
  title={Image quality metrics: PSNR vs. SSIM},
  author={Hore, Alain and Ziou, Djemel},
  booktitle={2010 20th international conference on pattern recognition},
  pages={2366--2369},
  year={2010},
  organization={IEEE}
}

@article{ssim,
  title={Image quality assessment: from error visibility to structural similarity},
  author={Wang, Zhou and Bovik, Alan C and Sheikh, Hamid R and Simoncelli, Eero P},
  journal={IEEE transactions on image processing},
  volume={13},
  number={4},
  pages={600--612},
  year={2004},
  publisher={IEEE}
}

@inproceedings{lpips,
  title={The unreasonable effectiveness of deep features as a perceptual metric},
  author={Zhang, Richard and Isola, Phillip and Efros, Alexei A and Shechtman, Eli and Wang, Oliver},
  booktitle={Proceedings of the IEEE conference on computer vision and pattern recognition},
  pages={586--595},
  year={2018}
}

@article{fid,
  title={Gans trained by a two time-scale update rule converge to a local nash equilibrium},
  author={Heusel, Martin and Ramsauer, Hubert and Unterthiner, Thomas and Nessler, Bernhard and Hochreiter, Sepp},
  journal={Advances in neural information processing systems},
  volume={30},
  year={2017}
}

@inproceedings{fid-vid,
  title={Conditional GAN with Discriminative Filter Generation for Text-to-Video Synthesis.},
  author={Balaji, Yogesh and Min, Martin Renqiang and Bai, Bing and Chellappa, Rama and Graf, Hans Peter},
  booktitle={IJCAI},
  volume={1},
  number={2019},
  pages={2},
  year={2019}
}

@article{fvd,
  title={Towards accurate generative models of video: A new metric \& challenges},
  author={Unterthiner, Thomas and Van Steenkiste, Sjoerd and Kurach, Karol and Marinier, Raphael and Michalski, Marcin and Gelly, Sylvain},
  journal={arXiv preprint arXiv:1812.01717},
  year={2018}
}

@inproceedings{mcld,
  title={Multi-focal Conditioned Latent Diffusion for Person Image Synthesis},
  author={Liu, Jiaqi and Zhang, Jichao and Rota, Paolo and Sebe, Nicu},
  booktitle={Proceedings of the Computer Vision and Pattern Recognition Conference},
  pages={16019--16028},
  year={2025}
}

@article{moviegen,
  title={Movie gen: A cast of media foundation models},
  author={Polyak, Adam and Zohar, Amit and Brown, Andrew and Tjandra, Andros and Sinha, Animesh and Lee, Ann and Vyas, Apoorv and Shi, Bowen and Ma, Chih-Yao and Chuang, Ching-Yao and others},
  journal={arXiv preprint arXiv:2410.13720},
  year={2024}
}

@article{phantom,
  title={Phantom: Subject-consistent video generation via cross-modal alignment},
  author={Liu, Lijie and Ma, Tianxiang and Li, Bingchuan and Chen, Zhuowei and Liu, Jiawei and Li, Gen and Zhou, Siyu and He, Qian and Wu, Xinglong},
  journal={arXiv preprint arXiv:2502.11079},
  year={2025}
}

@article{ipadapter,
  title={Ip-adapter: Text compatible image prompt adapter for text-to-image diffusion models},
  author={Ye, Hu and Zhang, Jun and Liu, Sibo and Han, Xiao and Yang, Wei},
  journal={arXiv preprint arXiv:2308.06721},
  year={2023}
}

@inproceedings{shi2025self,
  title={Self-supervised ControlNet with Spatio-Temporal Mamba for Real-world Video Super-resolution},
  author={Shi, Shijun and Xu, Jing and Lu, Lijing and Li, Zhihang and Hu, Kai},
  booktitle={Proceedings of the Computer Vision and Pattern Recognition Conference},
  pages={7385--7395},
  year={2025}
}

@inproceedings{dwpose,
  title={Effective whole-body pose estimation with two-stages distillation},
  author={Yang, Zhendong and Zeng, Ailing and Yuan, Chun and Li, Yu},
  booktitle={Proceedings of the IEEE/CVF International Conference on Computer Vision},
  pages={4210--4220},
  year={2023}
}

@misc{paddleocr2021,
  title={PaddleOCR: Awesome Multilingual OCR Toolkits},
  author={PaddlePaddle},
  year={2021},
  howpublished={\url{https://github.com/PaddlePaddle/PaddleOCR}}
}

@misc{pyiqa2022,
  title={IQA-PyTorch: PyTorch Toolbox for Image Quality Assessment},
  author={Chen, Chaofeng},
  year={2022},
  howpublished={\url{https://github.com/chaofengc/IQA-PyTorch}}
}

@inproceedings{omnihuman,
  title={Omnihuman-1: Rethinking the scaling-up of one-stage conditioned human animation models},
  author={Lin, Gaojie and Jiang, Jianwen and Yang, Jiaqi and Zheng, Zerong and Liang, Chao and Zhang, Yuan and Liu, Jingtuo},
  booktitle={Proceedings of the IEEE/CVF International Conference on Computer Vision},
  pages={13847--13858},
  year={2025}
}

@article{guo2024liveportrait,
  title   = {LivePortrait: Efficient Portrait Animation with Stitching and Retargeting Control},
  author  = {Guo, Jianzhu and Zhang, Dingyun and Liu, Xiaoqiang and Zhong, Zhizhou and Zhang, Yuan and Wan, Pengfei and Zhang, Di},
  journal = {arXiv preprint arXiv:2407.03168},
  year    = {2024}
}

@InProceedings{Siarohin_2019_NeurIPS,
  author={Siarohin, Aliaksandr and Lathuilière, Stéphane and Tulyakov, Sergey and Ricci, Elisa and Sebe, Nicu},
  title={First Order Motion Model for Image Animation},
  booktitle = {Conference on Neural Information Processing Systems (NeurIPS)},
  month = {December},
  year = {2019}
}

@article{SMIRK,
  title={SMIRK: 3D Facial Expressions through Analysis-by-Neural-Synthesis},
  author={Retsinas, George and Filntisis, Panagiotis P and Danecek, Radek and Abrevaya, Victoria F and Roussos, Anastasios and Bolkart, Timo and Maragos, Petros},
  journal={arXiv preprint arXiv:2404.04104},
  year={2024}
}

@inproceedings{deng2019arcface,
  title={Arcface: Additive angular margin loss for deep face recognition},
  author={Deng, Jiankang and Guo, Jia and Xue, Niannan and Zafeiriou, Stefanos},
  booktitle={CVPR},
  year={2019}
}
}

\clearpage
\appendix
\setcounter{page}{1}
\maketitlesupplementary

\newsavebox{\lefttable}
\renewcommand{\topfraction}{0.9}      
\renewcommand{\dbltopfraction}{0.9}   
\setcounter{topnumber}{2}             
\setcounter{dbltopnumber}{2}          
\begin{table*}[t]
\caption{Model architecture, training hyperparameters, and inference cost.}
\vspace{-0.3cm}
\centering
\renewcommand{\arraystretch}{1}
\label{tab:config}
\sbox{\lefttable}{%
\resizebox{0.56\textwidth}{!}{%
\begin{tabular}{l | c c }
\toprule
\textbf{Config} & \textbf{One-to-All-1.3B} & \textbf{One-to-All-14B} \\
\midrule
Hidden Dim & 1536   & 5120 \\
Num Layers -- Extractor & 14 & 7 \\
Num Layers -- DiT         & 30  & 40 \\
Pre-trained Model & Wan2.1-T2V-1.3B   & Wan2.1-T2V-14B \\
\midrule
\multicolumn{3}{l}{\textbf{Training Parameters}} \\
\midrule
Batch Size per GPU       & \multicolumn{2}{c}{1} \\
Optimizer                & \multicolumn{2}{c}{AdamW} \\
Learning Rate            & \multicolumn{2}{c}{$1\times10^{-5}$} \\
Weight Decay             & \multicolumn{2}{c}{$1\times10^{-4}$} \\
LR Schedule              & \multicolumn{2}{c}{constant with warmup} \\
Training Steps (Stage 1/2/3) & 40k / 40k / 20k  & 30k / 30k / 20k \\
\midrule
\multicolumn{3}{l}{\textbf{Inference Cost} ($81\times576\times1024$, 30steps, bfloat16, H100)} \\
\midrule
Memory (GB)          & 29 & 65  \\
Time (mm:ss, $\lambda_P = \lambda_I = 1.0$) & 0:59 & 3:41\\ 
Time (mm:ss, $\lambda_P = \lambda_I > 1.0$) & 1:58 & 7:43  \\
Time (mm:ss, $\lambda_P \neq \lambda_I > 1.0$) & 2:57 & 11:12\\
\bottomrule
\end{tabular}%
}%
}
\noindent
\begin{minipage}[t]{0.5\textwidth}
\centering
\usebox{\lefttable}
\end{minipage}%
\hfill
\begin{minipage}[t]{0.42\textwidth}
\centering
\resizebox{!}{\ht\lefttable}{%
\begin{tabular}{l c}
\toprule
\multicolumn{2}{l}{\textbf{PoseResNet}} \\
\midrule
Conv3d & 3$\to$16, kernel=3, stride=1 \\
ResNet Block 1 & -- \\
Conv3d & 16$\to$16, kernel=3, stride=(1,2,2) \\
ResNet Block 2 & -- \\
Conv3d & 16$\to$16, kernel=3, stride=(2,2,2) \\
ResNet Block 3 & -- \\
Conv3d & 16$\to$16, kernel=3, stride=(2,2,2) \\
ResNet Block 4 & -- \\
Conv3d & 16$\to$16, kernel=3, stride=(1,2,2) \\
Conv3d & 16$\to$1536/5120, kernel=3, stride=1 \\
\midrule
\multicolumn{2}{l}{\textbf{ResNet Block}} \\
\midrule
GroupNorm & 4 groups, eps=1e-6 \\
SiLU & -- \\
Conv3d & 16$\to$16, kernel=3, stride=1 \\
GroupNorm & 4 groups, eps=1e-6 \\
SiLU & -- \\
Conv3d & 16$\to$16, kernel=3, stride=1 \\
\bottomrule
\end{tabular}%
}
\end{minipage}
\vspace{-0.4cm}
\end{table*}

\section{Details of Cartoon Dataset}
\begin{figure}[t]
    \centering
    \captionsetup[subfigure]{font=small, labelfont=footnotesize}
    
    \begin{subfigure}[b]{1\linewidth}
        \centering
        \includegraphics[width=\linewidth]{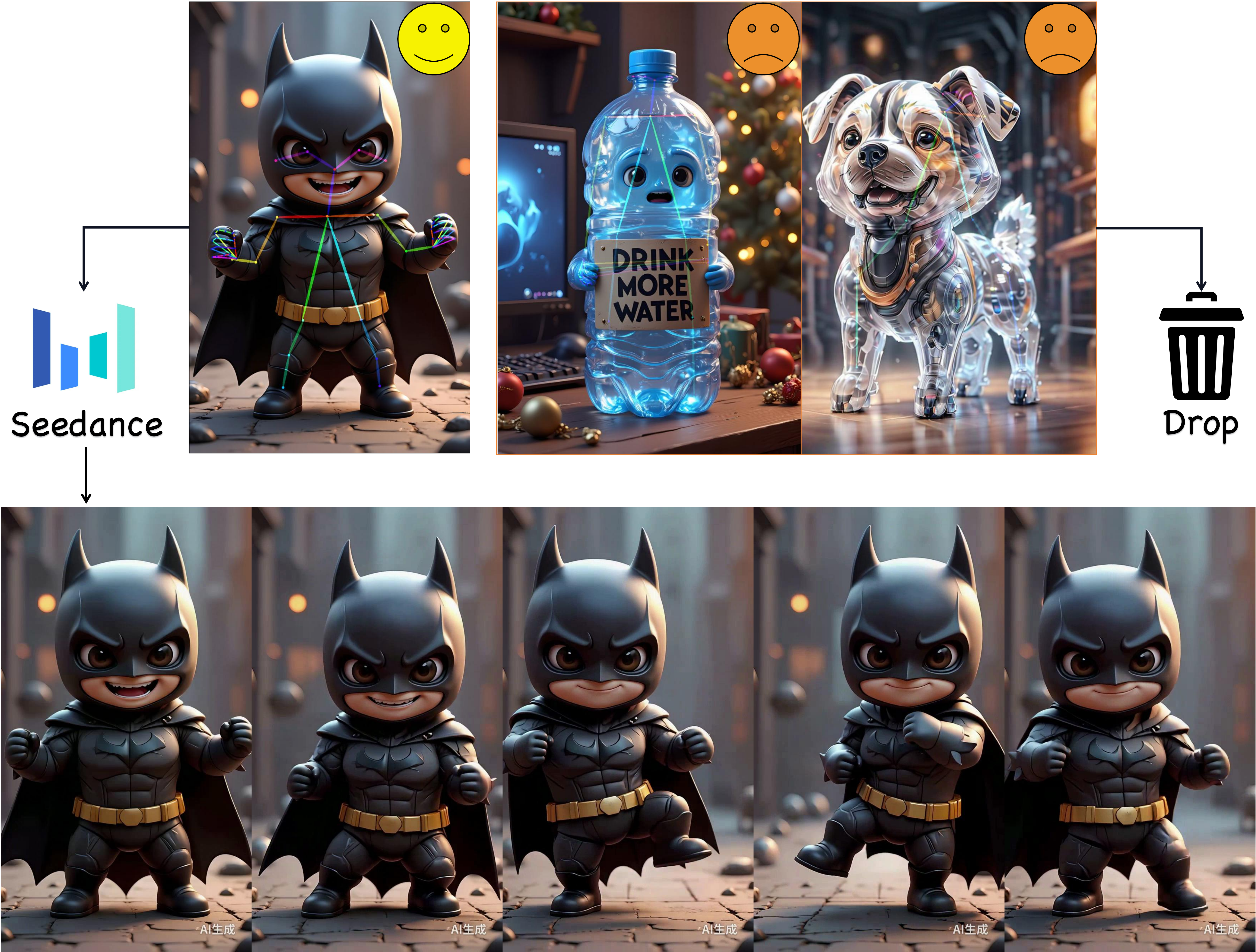}
        \caption{Dataset curation pipeline.}
        \label{fig:cartoon_pipeline}
    \end{subfigure}
    
    \vspace{0.1cm}
    
    \begin{subfigure}[b]{1\linewidth}
        \centering
        \includegraphics[width=\linewidth]{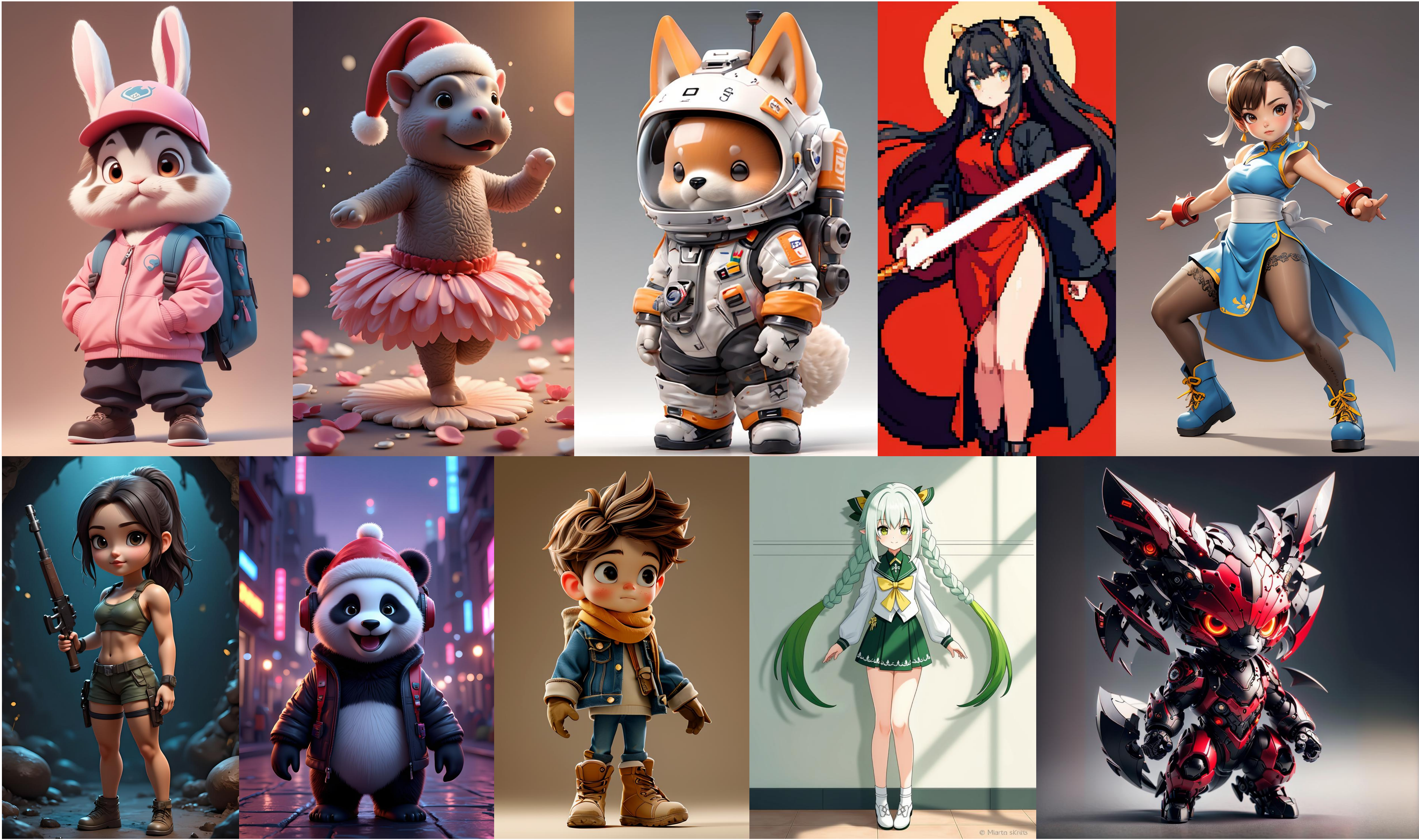}
        \caption{Examples from Cartoon dataset.}
        \label{fig:cartoon_examples}
    \end{subfigure}
    
    \vspace{-0.2cm}
    \caption{Cartoon Dataset construction: (a) pose filtering and video generation pipeline, (b) representative samples.}
    \label{fig:cartoon_dataset}
    \vspace{-0.3cm} 
\end{figure}

To enhance cross-style generalization, we construct a specialized cartoon and anime dataset.
We collect 1000 images from the AI art platform Shakker\footnote{\url{https://www.shakker.ai/}}, which provides diverse artistic styles and character designs. To ensure reliable pose conditioning, we manually filter out images where DWPose fails to detect clear body keypoints, resulting in 212 high-quality samples. 
We split these into 200 for training and 12 for evaluation. 
The overall curation process and representative examples are shown in Fig.~\ref{fig:cartoon_dataset}.
For each cartoon image, we generate a corresponding video clip using Seedance~\cite{seedance} with the text prompt ``a character is dancing''. Since Seedance outputs contain watermarks, we detect these regions and assign them zero loss weight during training, implementation details in Sec.~\ref{sec:region_loss}.

\section{Implementation Details}

\subsection{Training Details}
\noindent\textbf{Multi-resolution training.} 
We adopt a bucket-based sampler to support varying resolutions and aspect ratios training.
Each bucket is defined by a target resolution (e.g., 512px or 768px) and a set of aspect ratios. During sampling, we first select a resolution according to predefined probabilities, then match the sample to the closest aspect ratio within that bucket. Frames are center-cropped and resized to the target dimensions. For example, a $720 \times 1300$ video assigned to the 512px bucket would be resized to $384 \times 672$ (closest to $9{:}16$ aspect ratio). 
Images are treated as single-frame videos and processed identically.
To maintain a specific video-to-image data ratio (6:1 in our experiments) per epoch, we partition images into $K/6$ groups by aspect ratio, where $K$ is the number of video samples. When sampling images, we first select one group and then randomly pick an image from it.

\begin{figure}[t]
    \centering
    \includegraphics[width=\linewidth]{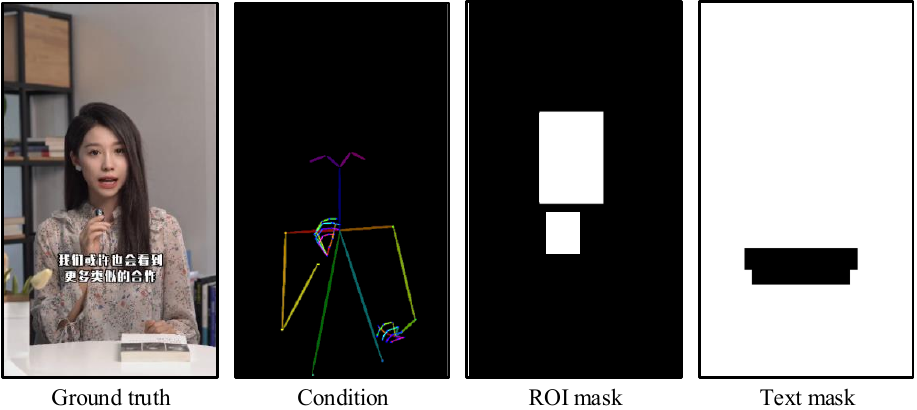}
    \vspace{-0.5cm}
    \caption{Visualization of region-weighted loss components.}
    \label{fig:region_mask}
    \vspace{-0.3cm}
\end{figure}

\noindent\textbf{Region-weighted loss.} 
\label{sec:region_loss}
To improve generation quality, we apply spatially adaptive loss weights during training. We partition each frame into regions of interest (ROI) and non-ROI.
ROI includes head regions and high-confidence hand regions, while non-ROI includes text regions. 
We extend the base rectified flow loss $\mathcal{L}_\mathrm{RF}$ (Eq.~\ref{eq:diffusion_loss}) with spatial weighting:
\begin{equation}
\mathcal{L}_{\mathrm{weighted}} = \mathcal{L}_{\mathrm{RF}} \cdot \mathcal{M}^{\mathrm{text}} \cdot \left( (w_{\mathrm{roi}} - 1) \cdot \mathcal{M}^{\mathrm{roi}} + 1 \right),
\label{eq:region_weighted_loss}
\end{equation}
where $\mathcal{M}^{\mathrm{text}} \in \{0,1\}$ is the inverse text mask (0 for text regions, 1 otherwise), $\mathcal{M}^{\mathrm{roi}} \in \{0,1\}$ denotes the binary ROI mask, and $w_{\mathrm{roi}}$ is the ROI weight usually set 3.0.

In implementation, we use DWPose~\cite{dwpose} for body detection and PaddleOCR~\cite{paddleocr2021} for text detection. 
Hand regions are treated as ROI only when all corresponding keypoint confidences exceed 0.8. 
The resulting masks are spatially downsampled by $16\times$ and temporally merged over every 4 consecutive frames to match the latent's feature dimension. Fig.~\ref{fig:region_mask} visualizes example region masks.

\subsection{Evaluation Details}
\noindent\textbf{Metrics.}
We quantitatively evaluate our results using several metrics, including PSNR, SSIM, LPIPS, FID, FID-VID and FVD. For misaligned-scenario evaluation, we adopt three metrics following LivePortrait~\cite{guo2024liveportrait}: CSIM, AED and APD-body. The detailed metrics are introduced as follows:
\begin{itemize}[leftmargin=*]
    \item \textbf{PSNR} quantifies the pixel-level reconstruction accuracy by measuring the ratio between maximum signal power and noise power, expressed in decibels (dB). Higher PSNR values indicate better reconstruction quality.
    
    \item \textbf{SSIM} evaluates image similarity by comparing luminance, contrast, and structural patterns. Unlike PSNR, SSIM better reflects human perception of image quality.
    
    \item \textbf{LPIPS} measures perceptual distance using deep features extracted from pretrained networks. It correlates more closely with human judgment than traditional metrics.
    
    \item \textbf{FID} computes the distance between feature distributions of real and generated images using Inception network embeddings. Lower FID indicates generated images are closer to real data in feature space.
    
    \item \textbf{FID-VID} extends FID to videos by computing distribution distance on frame-level features, capturing temporal consistency.
    
    \item \textbf{FVD} measures video quality by comparing feature distributions that encode both spatial appearance and temporal dynamics. Lower values indicate better video fidelity.
    
    \item \textbf{CSIM} measures identity preservation by computing the cosine similarity between ArcFace~\cite{deng2019arcface} embeddings of the source and generated faces. Higher values indicate better identity consistency.

    \item \textbf{AED} evaluates expression transfer accuracy by computing the average L1 distance between 3DMM expression coefficients extracted by SMIRK~\cite{SMIRK} of the generated and driving faces. Lower values indicate more faithful expression reproduction.

    \item \textbf{APD-body} measures body-pose alignment by computing the average L1 distance between body keypoints of the generated and driving frames, extracted by DWPose~\cite{dwpose} with head keypoints excluded. Lower values indicate better alignment.
    
\end{itemize}

\noindent\textbf{Implementation.}
For video datasets, we follow the evaluation codebase from DisCo~\cite{disco}. For image datasets, we use the PyIQA~\cite{pyiqa2022} library for metric computation. Specifically, we use LPIPS-VGG for all LPIPS measurements.
All evaluations are performed on a single NVIDIA H100 GPU. We will release the full inference and evaluation code to facilitate reproduction.

\subsection{Hyperparameters}
Table~\ref{tab:config} summarizes the model architecture, training hyperparameters, and inference cost. The table also details the PoseResNet architecture, which applies the same spatial-temporal compression as the VAE encoder. Its final layer projecting features to match the backbone dimension for element-wise addition.
For training, we use the AdamW optimizer with a learning rate of $1\times 10^{-5}$ and weight decay of $1\times 10^{-4}$ following a constant schedule with warmup.
The three-stage training runs for 40k / 40k / 20k steps for the 1.3B model and 30k / 30k / 20k steps for the 14B model. 
For inference, we benchmark both models on a single H100 GPU to generate 81 frames at $576 \times 1024$ resolution. Under our default guidance setting ($\lambda_P = \lambda_I = 1.5$), the 1.3B model requires 29\,GB of memory with a runtime of 1\,min\,58\,s, while the 14B model requires 65\,GB with a runtime of 7\,min\,43\,s. These numbers are reported without any compression or acceleration techniques and can be further optimized to support lower‑memory devices.

\section{Additional Results}
\subsection{More Ablation Studies}
\noindent\textbf{Ablation on RoPE Design in HRFA.}
A key design of HRFA is applying 3D RoPE with $f=0$ to the cross-attention query. Here we provide additional ablation results to justify this design.
\begin{figure}[t]
    \centering
    \includegraphics[width=\linewidth]{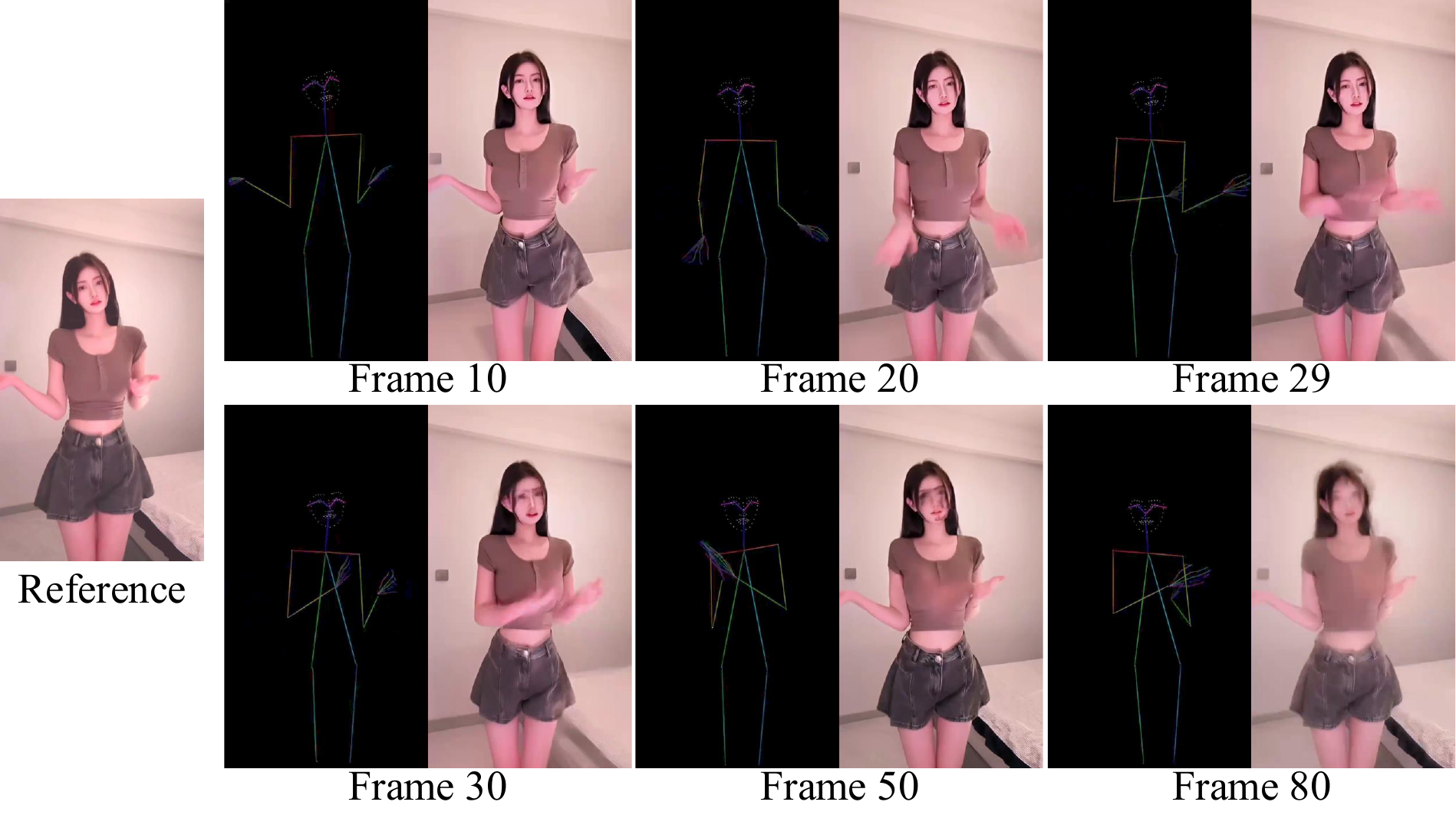}
    \vspace{-0.4cm}
    \caption{Failure case of retaining full 3D RoPE.}
    \label{fig:ablation_rope}
    \vspace{-0.2cm}
\end{figure}
We train a variant that retains full 3D RoPE encoding for $\mathrm{Q}'$ (computed as $\mathrm{RoPE}_{3D}(h W_q)$ without setting $f=0$) on 29-frame video clips. During inference with 81-frame sequences, this variant maintains reasonable quality for the first 29 frames but exhibits severe collapse afterward, as shown in Fig.~\ref{fig:ablation_rope}. This result indicates that the model overfits to absolute frame positions and cannot extrapolate beyond the training sequence length.
In contrast, our design with $f=0$ preserves the model's temporal extrapolation capability and avoids such degradation.

\begin{table}[t]
\centering
\caption{Ablation on training stage design. All methods evaluated on TikTok benchmark using 14B model.}
\vspace{-0.2cm}
\label{tab:train_stage_ablation}
\resizebox{0.85\linewidth}{!}{%
\begin{tabular}{l c c c}
\toprule
\textbf{Training Strategy} & \textbf{SSIM$\uparrow$} & \textbf{LPIPS$\downarrow$} & \textbf{FVD$\downarrow$} \\
\midrule
Pose $\rightarrow$ Ref+Pose & 0.732 & 0.332 & 408.7 \\
Joint (Ref+Pose) & 0.729 & 0.343 & 415.3 \\
Ref $\rightarrow$ Ref+Pose (Ours) & \textbf{0.773} & \textbf{0.280} & \textbf{355.2} \\
\bottomrule
\end{tabular}}
\vspace{-0.3cm}
\end{table}

\noindent\textbf{Ablation on Training Stage.}
As described in Sec.~\ref{sec:three_stage_training}, we employ a two-stage training before token replace: stage~1 trains the reference extractor with reference condition only, and stage~2 adds pose conditioning. To validate this design, we compare three training strategies: (1)~Ref~$\rightarrow$~Ref+Pose (Ours), (2)~Pose~$\rightarrow$~Ref+Pose, and (3)~Joint training from scratch. All variants exclude face region enhancement and reference-guided pose control for fair comparison.
Table~\ref{tab:train_stage_ablation} shows our approach outperforms both alternatives across all metrics. We observe that introducing pose conditioning too early causes the model to over-rely on pose signals for both structure and appearance, making it difficult to learn robust identity features. This is consistent with findings from OmniHuman~\cite{omnihuman}, which also found that introducing stronger conditions (pose) too early can hinder the learning of weaker conditions (appearance). Our staged design addresses this by training appearance first, then layering motion control on top.

\begin{figure}[t]
    \centering
    \includegraphics[width=\linewidth]{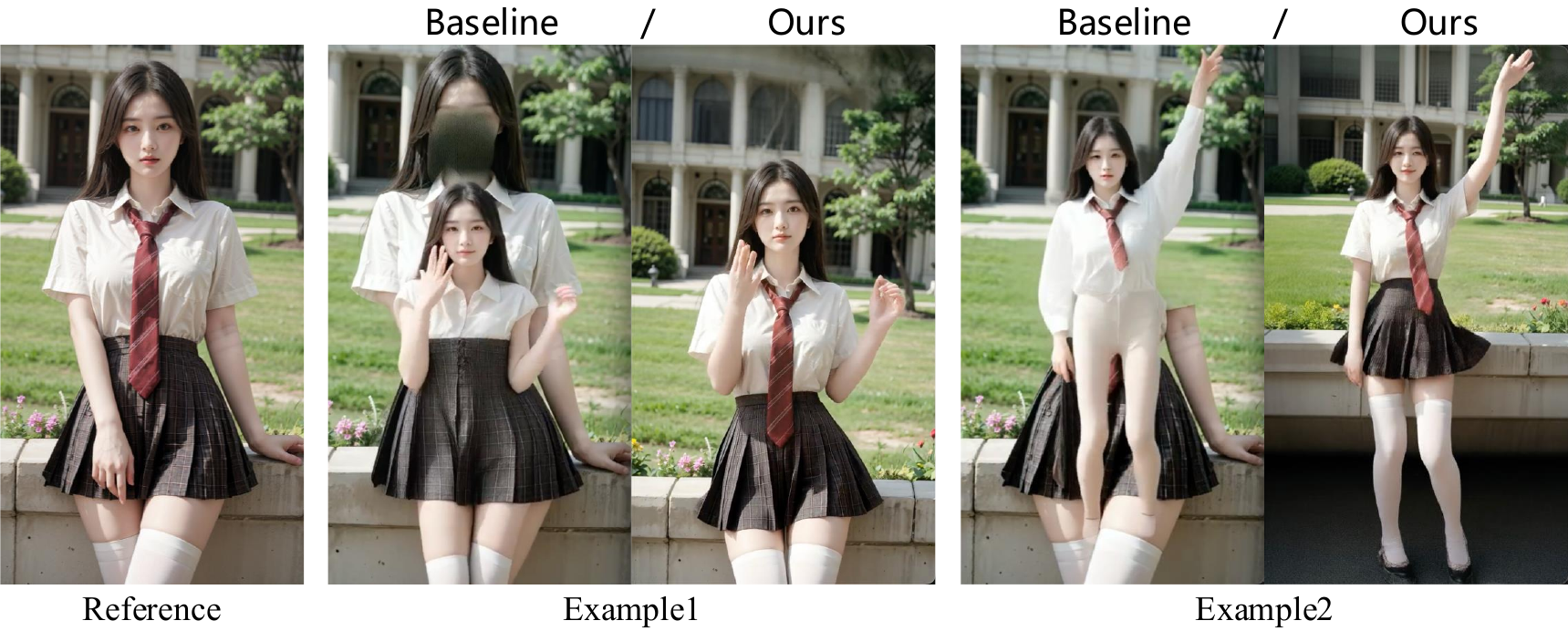}
    \vspace{-0.6cm}
    \caption{Failure cases of crop-and-resize baseline.}
    \label{fig:ablation_outpainting}
    \vspace{-0.5cm}
\end{figure}

\noindent\textbf{Ablation on outpainting strategy.}
To validate our self-supervised outpainting strategy, we compare it with a baseline that applies random crop-and-resize augmentation in training stage~2. As shown in Fig.~\ref{fig:ablation_outpainting}, the baseline produces severe artifacts due to spatial misalignment, which we attribute to conflicting training goals.
In stage~1, the model learns self-reconstruction with only the reference image as condition. Without pose guidance, pixel-level alignment between the reference and output is essential for effective learning. Introducing crop-and-resize augmentation at such stage would break the alignment. Applying it only in stage~2 creates a distribution shift: the reconstruction prior learned in stage~1 becomes invalid, forcing the model to relearn spatial mappings from scratch and resulting in training instability and blur.

In contrast, our masking-based approach preserves the spatial alignment of visible regions and enables outpainting training from the beginning, allowing seamless transfer of the reconstruction prior to pose-guided generation.

\section{More Results}
\begin{figure*}[t]
    \centering
    \includegraphics[width=\linewidth]{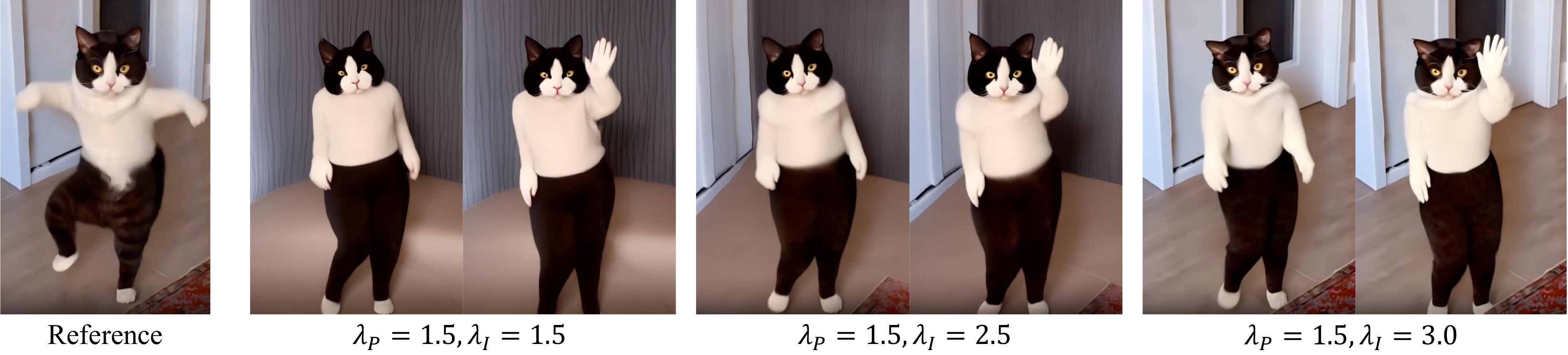}
    \vspace{-0.6cm}
    \caption{Results from the 1.3B model showing that higher $\lambda_I$ improves character and background consistency.
    }
    \label{fig:cfg}
    \vspace{0.2cm}
    \includegraphics[width=\linewidth]{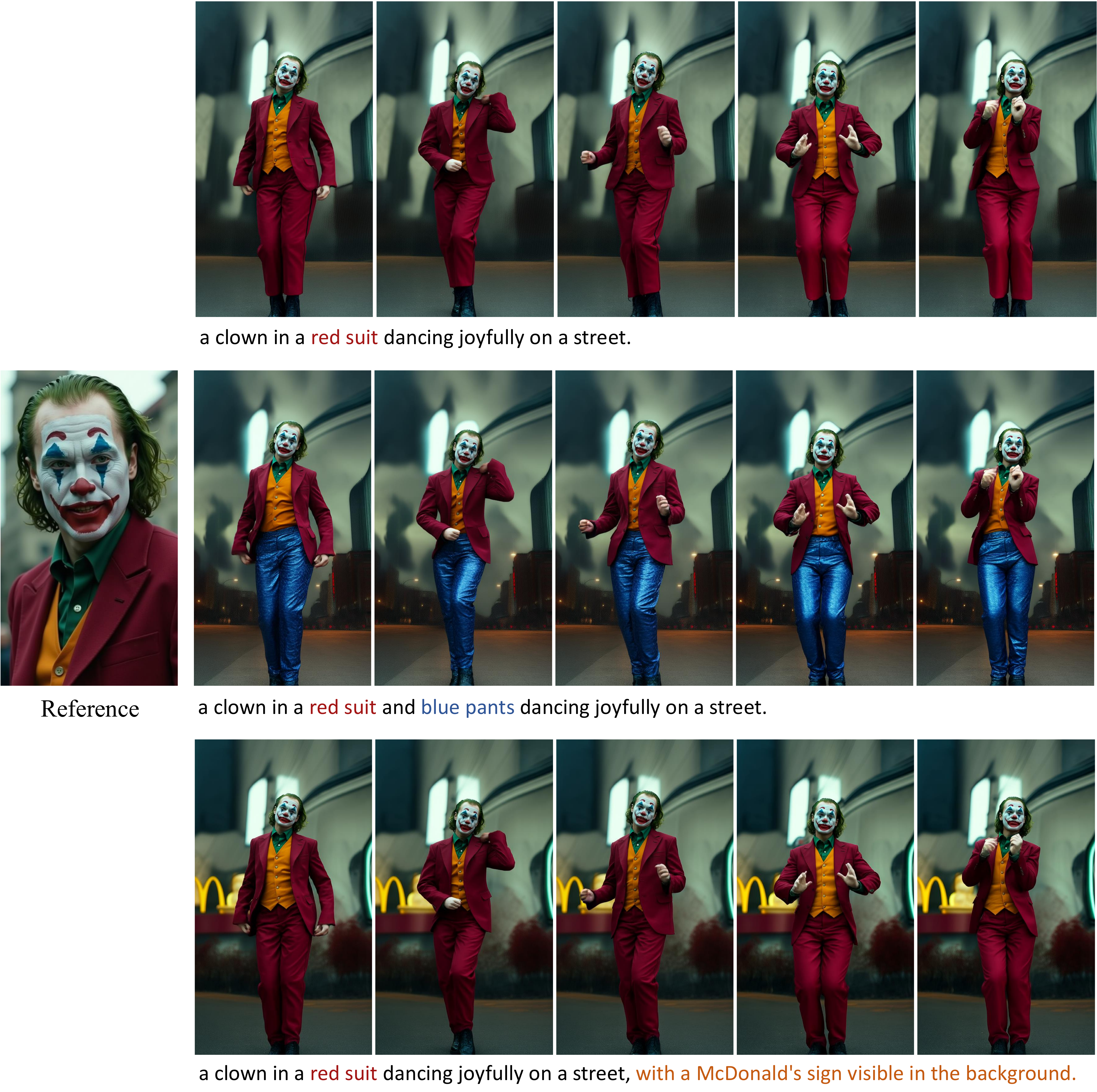}
    \vspace{-0.5cm}
    \caption{
    Our model enables prompt-based editing while maintaining identity and motion.
    }
    \vspace{-0.2cm}
    \label{fig:text_editing}
\end{figure*}

\begin{figure*}[t]
    \centering
    \includegraphics[width=\linewidth,height=\textheight,keepaspectratio]{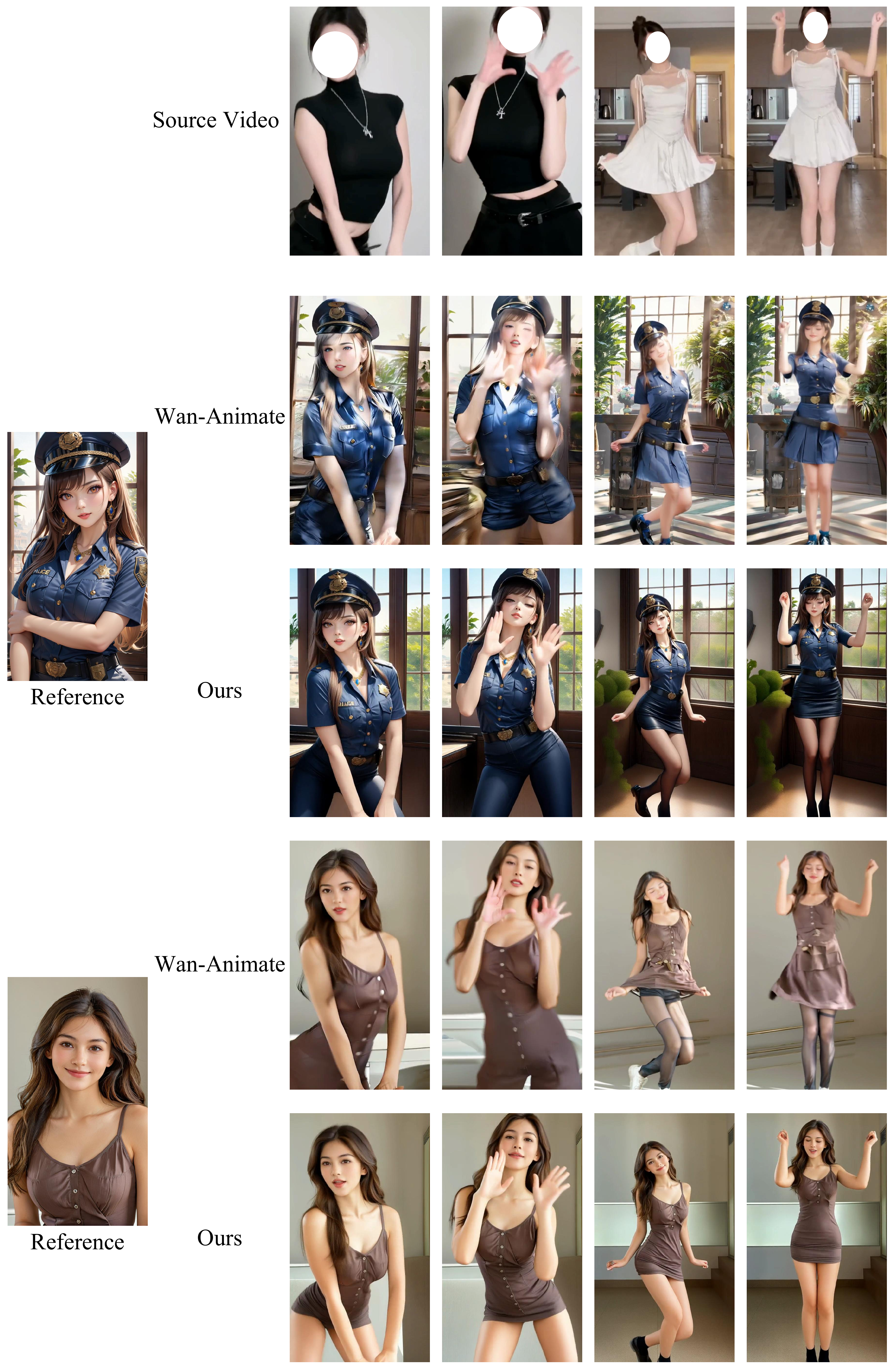}
    \caption{Visual comparison with Wan-Animate~\cite{wananimate} on misaligned image-video pairs (Part 1). Our method achieves better identity preservation and fewer visual artifacts.}
    \label{fig:misaligned_comparison1}
\end{figure*}

\begin{figure*}[t]
    \centering
    \includegraphics[width=\linewidth,height=\textheight,keepaspectratio]{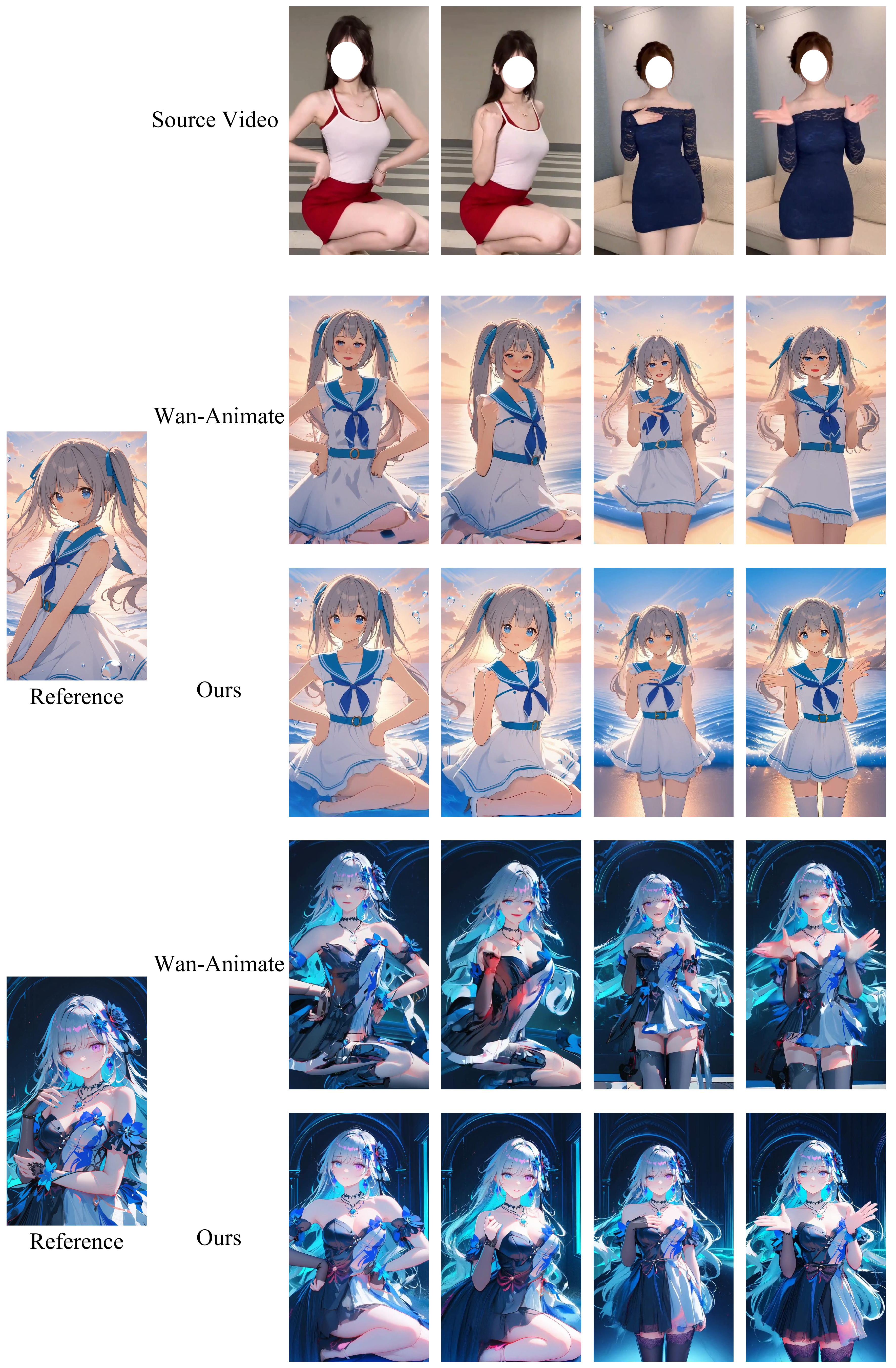}
    \caption{Visual comparison with Wan-Animate~\cite{wananimate} on misaligned image-video pairs (Part 2). Our method achieves better identity preservation and fewer visual artifacts.}
    \label{fig:misaligned_comparison2}
\end{figure*}

\begin{figure*}[t]
    \centering
    \includegraphics[width=\linewidth]{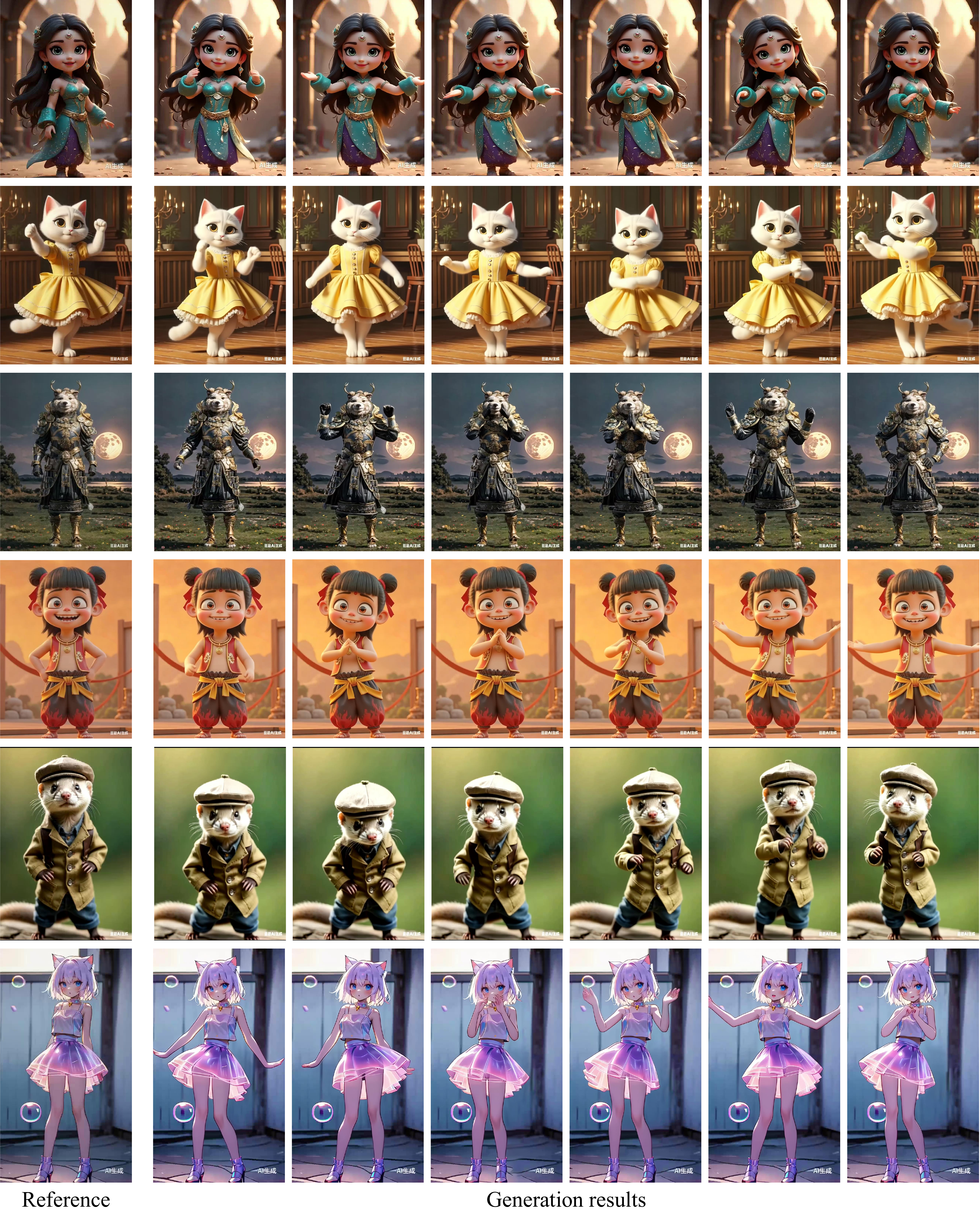}
    \caption{Additional Cartoon benchmark results.
    }
    \label{fig:cartoon_results}
\end{figure*}

We provide additional visualizations in Fig.~\ref{fig:cfg}, Fig.~\ref{fig:text_editing}, Fig.~\ref{fig:misaligned_comparison1}, Fig.~\ref{fig:misaligned_comparison2}, and Fig.~\ref{fig:cartoon_results}.
Fig.~\ref{fig:cfg} visualizes the performance of the 1.3B model under different guidance settings. We find that increasing the image guidance scale $\lambda_I$ progressively improves character and background consistency, resulting in more robust performance on out-of-domain inputs.
Fig.~\ref{fig:text_editing} shows that our model supports text edits of unseen elements, such as pants or environment, while maintaining identity and motion.
Fig.~\ref{fig:misaligned_comparison1} and Fig.~\ref{fig:misaligned_comparison2} show qualitative comparisons with Wan-Animate~\cite{wananimate} under misaligned settings.
Fig.~\ref{fig:cartoon_results} presents additional qualitative results on Cartoon benchmark.


\section{Limitations and Future Work}
Despite strong performance, our method has several limitations. 
First, we have not fully explored the optimal ratio between image and video training data. Our 1.3B model uses different checkpoints for image and video benchmarks, with the former fine-tuning on a higher image sampling ratio. 
Second, high-resolution or long-duration generation with the 14B model remains computationally expensive, requiring 65GB memory and over 7 minutes per inference. Future work could explore more efficient architectures or quantization methods to reduce this cost.
Finally, our method currently relies solely on pose sequences, which only implicitly capture camera motion through character positioning. This limits precise control over camera trajectories independent of character motion. Future work could introduce explicit camera parameters or motion cues to enable independent control over camera movement.

\end{document}